%% file: main.tex
\documentclass{article}
\usepackage{pact_arxiv,times}
\usepackage{xcolor}
\usepackage{booktabs}  
\usepackage[table]{xcolor}
\usepackage{booktabs,pifont}
\usepackage{multirow}
\usepackage{makecell}
\usepackage{amsmath,amsfonts,bm}
\usepackage{graphicx, booktabs, pifont, caption, multirow, xcolor}
\definecolor{okgreen}{HTML}{1B7F3B}
\definecolor{nored}{HTML}{B3261E}

\usepackage{hyperref}
\usepackage{url}
\usepackage{amssymb}
\usepackage{algorithm}
\usepackage{algpseudocode} 
\newcommand{\method}{PACT}
\definecolor{lightgreen}{RGB}{231, 245, 231}
\usepackage{wrapfig}

\usepackage[most]{tcolorbox}
\newcommand{\xmark}{\ding{55}}

\title{\method: Privileged Trace Co-Training for Multi-Turn Tool-Use Agents}

\author{\textbf{Zhenbang Du$^{1}$\thanks{Equal Contribution.}, Jun Luo$^{2}$\footnotemark[1], Zhiwei Zheng$^{3}$, Xiangchi Yuan$^{1}$, Kejing Xia$^{1}$} \\[2pt]
\textbf{Dachuan Shi$^{1}$, Qirui Jin$^{1}$, Qijia He$^{2}$, Shaofeng Zou$^{4}$, Yingbin Liang$^{2}$, Wenke Lee$^{1}$} \\[5pt]
{\normalfont $^{1}$Georgia Institute of Technology, $^{2}$Ohio State University} \\[1pt]
{\normalfont $^{3}$University of Pennsylvania, $^{4}$Arizona State University}
}

\iclrfinalcopy
\hypersetup{
  pdftitle={PACT: Privileged Trace Co-Training for Multi-Turn Tool-Use Agents},
  pdfauthor={Zhenbang Du, Jun Luo, Zhiwei Zheng, Xiangchi Yuan, Kejing Xia, Dachuan Shi, Qirui Jin, Qijia He, Shaofeng Zou, Yingbin Liang, Wenke Lee},
  pdfborder={0 0 0}
}
\begin{document}
\raggedbottom

\maketitle

\begin{abstract}
Multi-turn tool-use agents must reason, call tools, and adapt to observations across several interaction turns.
Post-training such agents is challenging, as reinforcement learning often suffers from sparse rewards and weak credit assignment despite matching the prompt-only inference setting, while supervised fine-tuning on expert traces provides dense process supervision but can over-constrain the model to fixed trajectories.
To tackle this, we propose \textbf{\method{}}, a \textbf{P}rivileged tr\textbf{A}ce \textbf{C}o-\textbf{T}raining framework for multi-turn tool-use agents.
The key idea is to use expert traces only as training-time optimization signals rather than rollout-time hints.
\method{} keeps rollout generation prompt-only, then uses expert traces to guide optimization through two complementary signals: a trace-conditioned RL surrogate that evaluates prompt-only rollouts under expert-trace context, and a component-aware SFT loss that provides annealed process supervision over reasoning and tool-call components.
To reduce over-reliance on the training-only trace context, \method{} incorporates prompt-only anchoring with standard prompt-only RL updates.
Experiments on FTRL, BFCL, and ToolHop show that PACT consistently outperforms strong SFT- and RL-based baselines. Further analyses support the complementarity of the two training objectives and the benefits of component-aware supervision and prompt-only anchoring. These results highlight the effectiveness of privileged trace co-training for multi-turn tool-use learning.
The project page is available at \url{https://zhenbangdu.github.io/pact-project-page}.
\end{abstract}

\addtocontents{toc}{\protect\setcounter{tocdepth}{-1}}
\input{sections/1-intro}
\input{sections/2-related}

\input{sections/3-preliminaries}

\input{sections/4-method}
\input{sections/5-experiment}

\input{sections/6-conclusion}

\setlength{\bibsep}{2pt plus 0.5pt minus 0.5pt}
\bibliography{main}
\bibliographystyle{pact_arxiv}

\appendix
\addtocontents{toc}{\protect\setcounter{tocdepth}{2}}
\clearpage
\begingroup
\setcounter{tocdepth}{2}
\renewcommand{\contentsname}{Appendix Contents}
\hypersetup{pdfborder={0 0 0},linktoc=all}
\tableofcontents
\endgroup
\clearpage

\input{sections/7-appendix}

\end{document}

%% file: sections/1-intro.tex
\section{Introduction}
\label{sec:introduction}

Tool-use extends LLMs beyond static text generation by enabling interaction with external resources, including search engines~\citep{nakano2021webgpt,zheng2025deepresearcher,song2025r1,jin2025search},
computational tools~\citep{schick2023toolformer,gou2024tora,liao2024mario},
and user-defined APIs~\citep{li2023api,patil2024gorilla,xu2026envfactoryscalingtooluseagents,prabhakar2026apigen}. Many practical tool-use tasks are multi-turn, requiring an agent to reason about the current state, construct tool-calls through tool selection and
argument specification, incorporate returned observations, and decide whether to continue or produce a final answer~\citep{barres2025tau,wang2025ragen,wang2026ragen,qu2026matchtir}. Post-training such agents is therefore challenging because they require learning from final-task outcomes while also handling a multi-turn decision process.

Reinforcement Learning (RL) is a natural paradigm, which optimizes the policy from prompt-only rollouts using rewards based on task success or tool-use behavior~\citep{shao2024deepseekmath,yu2026dapo,qian2026toolrl, pang2025tic}.
This matches inference-time behavior, where no expert trace is available.
However, multi-turn tool-use rewards are often sparse or weakly localized, providing limited feedback on which intermediate decisions lead to success or failure~\citep{wei2025reinforcing,qu2026matchtir}. And it may struggle to discover successful multi-turn tool-use trajectories.
Expert traces provide complementary process supervision through successful tool-use trajectories~\citep{qin2024toolllm,song2024agentbank,chen2023fireact}.
However, Supervised Fine-tuning (SFT) on these traces learns from fixed targets and can over-constrain the model to a single reference trajectory, even when different tool-use trajectories could solve the same problem~\citep{chen2025atlas}.
Moreover, SFT alone does not explicitly assign credit to intermediate reasoning and tool-use decisions based on final task success~\citep{chen2025beyond}.

\begin{wrapfigure}{r}{0.56\textwidth}
    \vspace{-0.8\baselineskip}
    \centering
    \includegraphics[width=\linewidth]{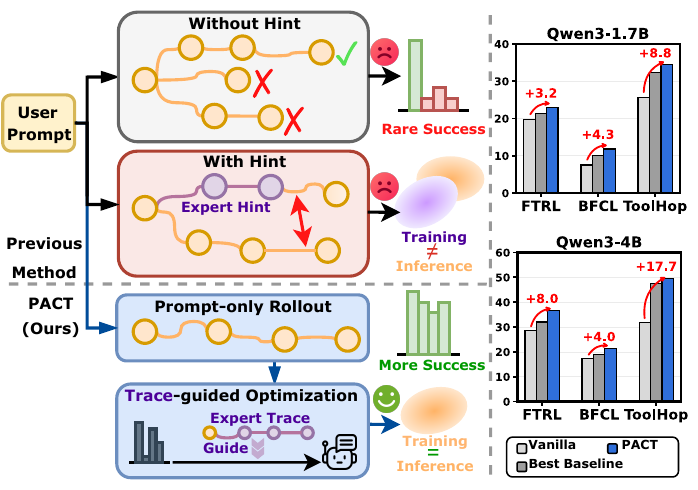}
    \caption{
    \method{} keeps rollouts prompt-only while using expert traces only as
    privileged optimization signals, improving multi-turn tool-use performance.
    FTRL reports Solve-F1.
    }
    \label{fig:compare}
    \vspace{-0.5\baselineskip}
\end{wrapfigure}

These limitations motivate methods that combine expert traces with RL, but existing combinations do not directly address this setting.
Hint-based methods provide expert solutions as additional rollout context, introducing a mismatch between hint-conditioned training rollouts and prompt-only inference~\citep{liu2025uft}.
Joint SFT-RL methods add imitation losses to RL, but are mainly designed for single-turn reasoning traces rather than long interactions with tool-calls and observations~\citep{chen2025beyond,yuan2025mitigating,zhang2026onpolicy}.
Taken together, these approaches either change the rollout conditioning or apply imitation objectives that are not specifically tailored to the structure of multi-turn tool-use behavior.
We therefore ask \emph{how expert traces can guide joint SFT--RL training for multi-turn tool-use agents while preserving prompt-only rollout}.

We answer this question with \textbf{\method{}}, a \textbf{P}rivileged tr\textbf{A}ce \textbf{C}o-\textbf{T}raining framework for multi-turn tool-use agents, as shown in Figure \ref{fig:compare}.
Instead of using expert traces to condition rollouts or treating them as full imitation targets, \method{} separates prompt-only rollout generation from trace-based optimization.
Within this framework, expert traces guide joint SFT--RL training by providing privileged context for RL updates and supervision targets for component-aware SFT.
Concretely, \method{} introduces two trace-based signals.
First, a trace-conditioned RL surrogate computes the policy likelihoods of prompt-only rollouts under expert-trace context, while retaining their original advantages for policy updates.
Second, a component-aware SFT loss supervises selected reasoning prefixes and complete tool-calls instead of the full trace.
With separate annealing schedules for different trace components, it provides early process guidance while gradually relaxing direct imitation as training progresses.
We also provide a conceptual latent-trace view to motivate the joint objective and interpret the two roles of expert traces in \method{} (Appendix~\ref{app:latent-trace-conceptual}).
To reduce over-reliance on the privileged trace context, \method{} further incorporates prompt-only anchoring, routing a subset of prompts to standard RL without trace-based loss in each optimization step.

Our contributions are summarized below.
\begingroup
\setlength{\leftmargini}{1.5em}
\begin{itemize}
    \setlength{\itemsep}{0.15em}
    \setlength{\parsep}{0pt}
    \setlength{\topsep}{0.2em}
    \setlength{\partopsep}{0pt}

\item We propose \method{}, a privileged trace co-training framework for multi-turn tool-use agents. \method{} separates prompt-only rollout generation from expert-trace-guided optimization.

\item We introduce two core designs in \method{}: trace-conditioned RL and component-aware SFT.  The RL surrogate uses expert traces as privileged context to guide trace-conditioned RL updates, while the SFT supervises selected reasoning prefixes and complete tool-call spans with component-aware annealing. 

\item Experiments on multiple multi-turn tool-use benchmarks show that \method{} consistently outperforms strong SFT- and RL-based baselines. Ablation studies support the empirical complementarity of trace-conditioned RL and component-aware SFT, as well as the benefits of component-aware supervision schedules.
\end{itemize}
\endgroup

%% file: sections/2-related.tex
\section{Related Work}
\label{sec:related_work}

\paragraph{Tool-use agents.}
Tool-use agents have been developed through prompting and instruction tuning to interleave reasoning with external actions, including ReAct, Toolformer, and ToolLLM~\citep{yao2022react,schick2023toolformer,qin2024toolllm}.
Recent work further applies RL to tool-use learning through interactive environments, reward design, and credit assignment.
ToolStar and FTRL construct environments for tool-use RL~\citep{dong2025toolstar,ye2025ftrl}, while Agent Lightning decouples agent execution from RL optimization~\citep{luo2025agent}.
ToolRL~\citep{qian2026toolrl} instead designs rewards for tool selection and application, and MatchTIR~\citep{qu2026matchtir} further assigns fine-grained credit to tool-integrated reasoning steps.
In contrast, \method{} does not modify the reward function or environment.
It introduces privileged trace co-training for multi-turn tool-use agents, using expert traces to jointly guide SFT and RL while preserving prompt-only rollout generation.

\paragraph{Combining SFT and RL.}
In general reasoning settings, recent work combines expert supervision with RL through different mechanisms.
UFT uses partial-solution hints to bridge imitation and exploration~\citep{liu2025uft}, BRIDGE coordinates SFT and RL through bilevel optimization~\citep{chen2025beyond}, and MIFO mitigates forgetting between the two training paradigms~\citep{yuan2025mitigating}.
CHORD combines on-policy RL with an auxiliary SFT objective controlled by a global coefficient and token-wise weights~\citep{zhang2026onpolicy}.
More recently, OPSD trains on student rollouts using token-level supervision from a teacher conditioned on privileged information~\citep{penaloza2026privileged,zhao2026opsd,yang2026RLSD}.
Multi-turn tool use, however, calls for supervision that accounts for the distinct roles of reasoning, tool-calls, and observations.
\method{} keeps rollout generation prompt-only and uses expert traces only during optimization, as privileged context for RL likelihood evaluation and process supervision over model-controllable trace components.

%% file: sections/3-preliminaries.tex
\section{Preliminaries}
\label{sec:preliminaries}

\subsection{Task Formulation}
\label{sec:task-formulation}

We consider multi-turn tool-use tasks, where an LLM agent is given a user prompt $x$ and a set of available tools $\mathcal{T}$, and interacts with an external tool environment over multiple turns. At each turn $t$, the agent generates a reasoning step $u_t$ and a structured tool-call $a_t$, which specifies an invoked tool from $\mathcal{T}$ and its arguments. The environment executes the tool-call and returns an observation $o_t$, which is appended to the interaction context. The process terminates when the agent produces a final answer $y$ or reaches a predefined maximum number of tool-use turns.

A rollout trajectory with $T$ tool-use turns can be written as
\begin{equation}
    \tau =
    \big(
    u_1,a_1,o_1,u_2,
    \ldots,
    u_T,a_T,o_T,
    y
    \big),
    \label{eq:rollout-trajectory}
\end{equation}
where $u_t$, $a_t$, and $y$ are generated by the agent, while $o_t$ is returned by the environment and used only as context for subsequent decisions. The rollout receives a scalar reward $R(x,\tau)$ that evaluates the final outcome and, when available, intermediate tool-use behavior.

In this work, we  additionally give each training prompt $x_i$  an expert tool-use trace
\begin{equation}
    h_i^* =
    \big(
    u^*_{i,1},a^*_{i,1},o^*_{i,1},
    \ldots,
    u^*_{i,T_i^*},a^*_{i,T_i^*},o^*_{i,T_i^*}
    \big).
    \label{eq:expert-trace}
\end{equation}
The detailed construction of expert traces is described in Appendix~\ref{app:training_data_construction}.
To avoid directly exposing the expert’s final answer, we exclude its explicit final-answer $y_i^*$ span from $h_i^*$.

\subsection{Post-training with SFT and RL}
\label{sec:post-training-sft-rl}

Given the rollout trajectories and expert traces defined above, two representative approaches for agentic LLM post-training are SFT and RL. Let $\pi_\theta$ denote the policy parameterized by $\theta$. SFT learns from expert traces, while RL optimizes the policy using rewards from rollouts.

\noindent\textbf{Supervised Fine-Tuning.}
Given an expert trace $h^*$ for prompt $x$, SFT maximizes the likelihood of the expert's model-generated tokens. Let $\mathcal{D}_{\mathrm{exp}}$ denote the expert-trace dataset, and let $\mathcal{M}(h^*)$ denote the model-generated token positions in the serialized trace. The SFT loss is given by
\begin{equation}
    \mathcal L_{\mathrm{SFT}}
    =
    -\mathbb E_{(x,h^*)\sim\mathcal D_{\mathrm{exp}}}
    \left[
    \sum_{n\in\mathcal M(h^*)}
    \log\pi_\theta\left(h_n^*\mid x,h_{<n}^*\right)
    \right].
\end{equation}
Tool observations remain in the conditioning history $h_{<n}^*$ but are excluded from the loss. This objective provides dense process supervision, but it trains the policy to follow fixed expert traces rather than optimizing the outcomes of its own rollouts. It can also over-constrain learning when multiple valid tool-use trajectories exist.

\noindent\textbf{Group-based Reinforcement Learning.}
One of the representative RL methods is Group Relative Policy Optimization (GRPO)~\citep{shao2024deepseekmath}, which estimates relative advantages from multiple rollouts sampled for the same prompt. For each prompt $x_i$, an old policy $\pi_{\mathrm{old}}$ samples a group of $K$ rollouts $\tau_{i,k} \sim \pi_{\mathrm{old}}(\cdot \mid x_i)$, $k=1,\ldots,K$. Each rollout receives a reward $R(x_i,\tau_{i,k})$, and the rewards are normalized within the group to obtain an advantage $A_{i,k}$, or its token-level variant $A_{i,k,n}$, where $n$ indexes positions of tokens generated in the rollout.

For a generated token $\tau_{i,k,n}$, the prompt-only importance ratio is
\begin{align}
    r^x_{i,k,n}(\theta)
    =\frac{\pi_\theta(\tau_{i,k,n}\mid x_i,\tau_{i,k,<n})}
    {\pi_{\mathrm{old}}(\tau_{i,k,n}\mid x_i,\tau_{i,k,<n})}.
\end{align}
The clipped GRPO loss with KL regularization is
\begin{equation}
    \mathcal L^x_{\mathrm{RL}}
    =-\mathbb E_{i,k,n}\!\left[
    \min\!\left\{r^x_{i,k,n}A_{i,k,n},\operatorname{clip}\!\left(r^x_{i,k,n},1-\epsilon,1+\epsilon\right)A_{i,k,n}\right\}
    -\beta_{\mathrm{KL}}D_{\mathrm{KL}}(\pi_\theta\|\pi_{\mathrm{ref}})
    \right],
\end{equation}
where $\epsilon$ is the clipping threshold, $\beta_{\mathrm{KL}}$ is the KL coefficient, and $\pi_{\mathrm{ref}}$ is a fixed reference policy. The expectation is over prompts, sampled rollouts, and model-generated tokens. This prompt-only formulation uses the task prompt and interaction history, without access to expert traces.

%% file: sections/4-method.tex
\begin{figure*}[t]
    \centering
    \includegraphics[width=\linewidth]{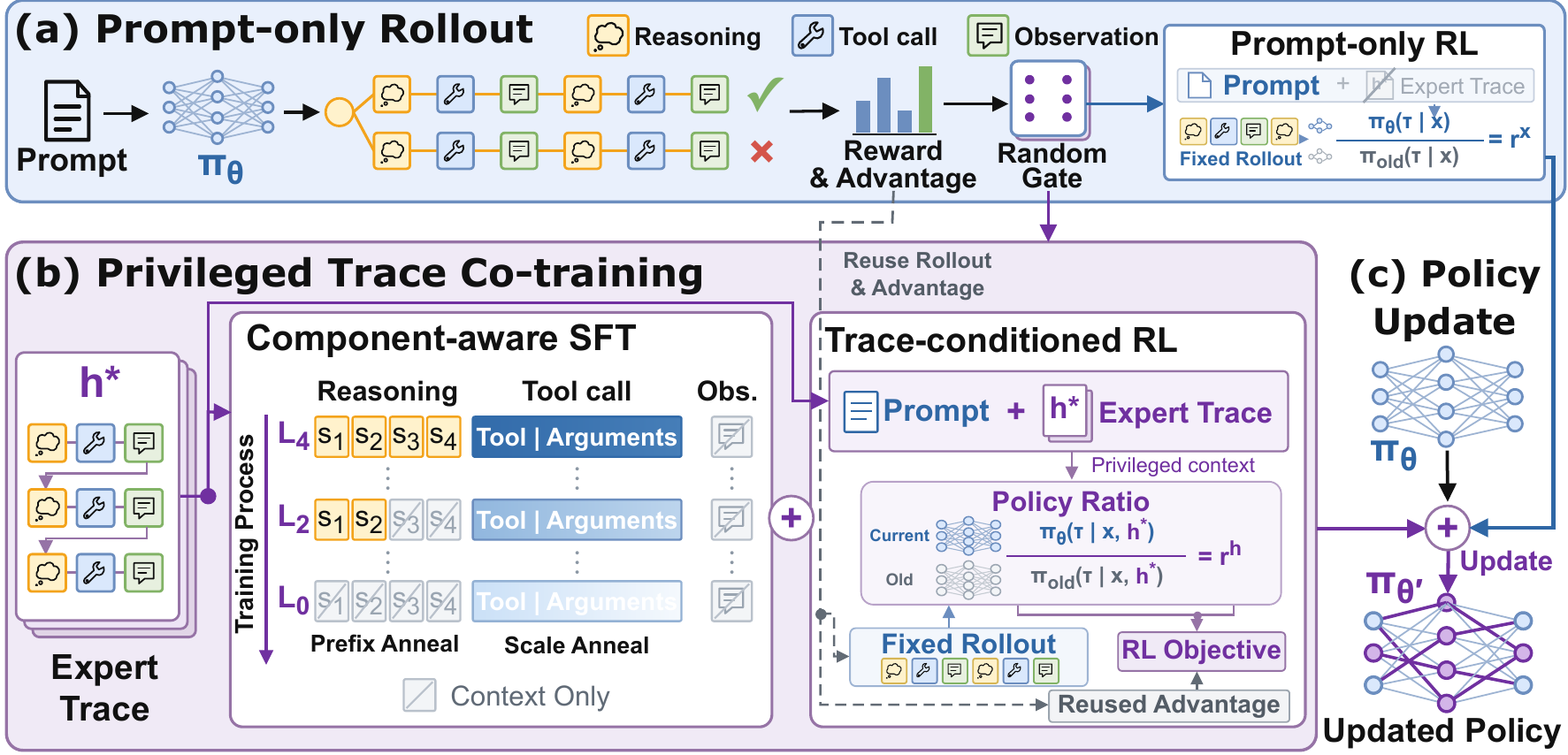}
    \caption{\textbf{Overview of \method{}.}
(a) Prompt-only rollouts provide rewards and advantages.
(b) Expert traces provide supervision targets for component-aware SFT and privileged context for trace-conditioned RL.
(c) The combined objectives update the policy.}
    \vspace{-10pt}
    \label{fig:overview}
\end{figure*}

\section{Method}
\label{sec:method}

In this section, we introduce \method{} as illustrated in Figure~\ref{fig:overview}.

\subsection{Prompt-only Rollout Generation}
\label{sec:method-rollout-generation}

For each training prompt \(x_i\), \method{} samples \(K\) tool-use rollouts from the old policy without access to the expert trace \(h_i^*\):
\begin{align}
\tau_{i,k}\sim\pi_{\mathrm{old}}(\cdot\mid x_i),
\qquad k=1,\ldots,K.
\end{align}
At each turn, the policy conditions on the task prompt and the accumulated interaction history, matching the inference-time input setting. Unlike hint-guided methods that provide partial expert solutions as rollout context \citep{liu2025uft}, \method{} uses expert traces only after rollout collection. Rewards and advantages are computed from the prompt-only rollouts, while expert traces provide privileged context for trace-conditioned RL and supervision targets for component-aware SFT.

\subsection{Trace-Conditioned RL Surrogate}
\label{sec:method-trace-rl}

After rollout generation, \method{} uses expert traces as privileged context for policy optimization. While rollout rewards and advantages determine which generated actions to reinforce, expert traces provide additional context for computing the token-level policy ratios used in the update.

For prompt $x_i$ with expert trace $h_i^*$, we define the privileged context as $c_i^h=[x_i;h_i^*]$ (see Appendix~\ref{app:training-settings} for details). We evaluate the same sampled rollout under both the current and old policies using this context, while keeping its rewards and advantages unchanged. Let $\mathcal{M}_{i,k}$ denote the model-generated token positions in rollout $\tau_{i,k}$. For $n\in\mathcal{M}_{i,k}$, the trace-conditioned policy ratio is
\begin{equation}
    r^h_{i,k,n}(\theta)
    =\frac{\pi_\theta(\tau_{i,k,n}\mid c_i^h,\tau_{i,k,<n})}
    {\pi_{\mathrm{old}}(\tau_{i,k,n}\mid c_i^h,\tau_{i,k,<n})}.
\label{eq:trace-conditioned-ratio}
\end{equation}
Using the token-level advantages $A_{i,k,n}$~\citep{qu2026matchtir} (Appendix~\ref{app:training_reward}), we optimize the following clipped trace-conditioned RL surrogate:
\begin{equation}
\begin{aligned}
\mathcal L^h_{\mathrm{RL}}(i)
= -\mathbb{E}_{k,\,n\in\mathcal{M}_{i,k}}\Big[
&\min\Big\{
 r^h_{i,k,n}A_{i,k,n},
 \operatorname{clip}\!\left(r^h_{i,k,n},1-\epsilon,1+\epsilon\right)A_{i,k,n}
\Big\}\\
&-\beta_{\mathrm{KL}}D_{\mathrm{KL}}(\pi_\theta\|\pi_{\mathrm{ref}})
\Big].
\end{aligned}
\label{eq:trace-rl-loss}
\end{equation}
This objective evaluates prompt-only rollouts under expert-trace context while retaining their original advantages. We treat the clipped policy objective as an auxiliary surrogate, rather than an importance correction between the sampling and evaluation contexts. 

\subsection{Component-aware SFT}
\label{sec:method-component-sft}

\begin{wrapfigure}{r}{0.40\textwidth}
    \vspace{-0.5\baselineskip}
    \centering
    \includegraphics[width=\linewidth]{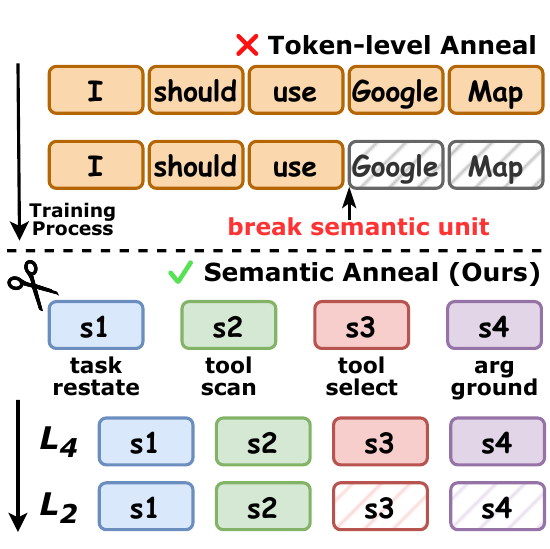}
    \caption{\textbf{Semantic reasoning-prefix annealing.} As training progresses, the supervised reasoning prefix is gradually shortened at semantic boundaries.}
    \label{fig:segment}
    \vspace{-0.5\baselineskip}
\end{wrapfigure}

Trace-conditioned RL uses expert traces as conditioning context to guide the policy update, but does not directly use them as supervision targets. 
We therefore add an auxiliary component-aware SFT loss that provides direct process supervision from expert traces. 
We supervise model-controllable components, \textit{reasoning tokens} and \textit{tool-call tokens}, while keeping tool observations only as context. 
Reasoning tokens guide intermediate decisions, but full reasoning imitation may over-constrain the policy. 
Tool-call tokens teach valid invocation, but exact teacher actions may represent only one of several valid choices. These differences motivate separate annealing strategies: we progressively shorten the supervised reasoning prefix within each assistant turn and lower the weight assigned to tool-call tokens while keeping their supervised spans complete.

\paragraph{Component-aware SFT loss.}
For prompt $x_i$, let $h^*_{i,n}$ denote the $n$-th token
in its expert trace. At training step $s$, we supervise
the active reasoning prefixes and complete tool-call spans,
with token positions indexed by $\mathcal R_i(s)$ and
$\mathcal A_i$, respectively. The corresponding binary
supervision masks are
\begin{equation}
    m^r_{i,n}(s)=\mathbf{1}[n\in\mathcal R_i(s)],
    \qquad
    m^a_{i,n}=\mathbf{1}[n\in\mathcal A_i].
\label{eq:sft-mask}
\end{equation}
We normalize the reasoning and tool-call losses separately and apply the tool-call weight after normalization:
\begin{equation}
\mathcal L_{\mathrm{SFT}}(s)
=-\sum_{i,n}\left[
\frac{m^r_{i,n}(s)}{\sum_{j,\ell}m^r_{j,\ell}(s)+\epsilon_{\mathrm{SFT}}}
+\lambda_a(s)\frac{m^a_{i,n}}{\sum_{j,\ell}m^a_{j,\ell}+\epsilon_{\mathrm{SFT}}}
\right]
\log\pi_\theta(h^*_{i,n}\mid x_i,h^*_{i,<n}).
\label{eq:sft-loss}
\end{equation}
Here, \(\epsilon_{\mathrm{SFT}}\) is a small constant for numerical stability. A component with no selected tokens contributes zero.

\paragraph{Annealed Reasoning Supervision.}
For reasoning, we supervise prefixes within each assistant turn to provide intermediate guidance without forcing imitation of the entire expert reasoning process.
Instead of annealing by raw token count, we construct prefix levels over semantic reasoning segments, as illustrated in Figure~\ref{fig:segment}.
Token-level annealing may end supervision in the middle of a coherent reasoning unit.
In contrast, segment-level annealing keeps the supervised prefixes aligned with semantic boundaries while gradually reducing supervision from longer to shorter prefixes, following $L_4 \rightarrow L_3 \rightarrow \cdots \rightarrow L_0$.
At $L_0$, no reasoning tokens receive direct SFT supervision, although they remain in the teacher-forced context.
Details of trace construction and the exact prefix-level definitions are provided in Appendix~\ref{app:training_data_construction}.

\paragraph{Annealed Tool-call Supervision.}
For tool-calls, we supervise complete spans so that tool names and all argument tokens receive direct supervision. Since an expert tool-call may be only one of several valid actions, we retain these spans throughout training while gradually reducing their imitation weight:
\begin{equation}
    \lambda_a(s)
    =\lambda_a^{\mathrm{start}}
    -\Delta\lambda_a\min\!\left(\frac{s}{S},1\right),
\label{eq:tool-call-annealing}
\end{equation}
where $\Delta\lambda_a=\lambda_a^{\mathrm{start}}-\lambda_a^{\mathrm{end}}$, $S$ is the annealing horizon, and $\lambda_a^{\mathrm{start}}>\lambda_a^{\mathrm{end}}$.

A larger early weight provides stronger supervision for complete tool-calls, while a smaller later weight relaxes imitation of the expert's specific tool choices and arguments.

\subsection{Overall Objective with Prompt-only Anchoring}
\label{sec:method-routing-objective}

For prompt $x_i$, \method{} combines the trace-conditioned RL surrogate with component-aware SFT to form the trace co-training objective:
\begin{equation}
\mathcal L_i^{\mathrm{trace}}
=\mathcal L_{\mathrm{RL}}^h(i)
+\beta_{\mathrm{SFT}}\mathcal L_{\mathrm{SFT}}(i;s),
\label{eq:trace-cotraining-objective}
\end{equation}
where $\mathcal L_{\mathrm{SFT}}(i;s)$ denotes the SFT contribution associated with the expert trace for $x_i$, and $\beta_{\mathrm{SFT}}$ controls its weight.

Expert traces play two related roles in this joint objective: they provide direct supervision for learning the expert process and privileged context for reward-based updates on self-generated rollouts. Appendix~\ref{app:latent-trace-conceptual} develops a conceptual latent-trace interpretation of these roles, while Section~\ref{sec training_objective} evaluates their empirical complementarity.

Although rollout generation is prompt-only, the trace-conditioned RL surrogate evaluates policy ratios under privileged trace context. To reduce over-reliance on this training-only context, \method{} also includes prompt-only anchoring.
For each prompt $x_i$, we sample a gate $z_i\sim\mathrm{Bernoulli}(q)$, shared by all $K$ rollouts, where $q$ is the prompt-only anchoring probability.

The final per-prompt objective selects between trace co-training and prompt-only RL:
\begin{equation}
\mathcal L_i
=(1-z_i)\mathcal L_i^{\mathrm{trace}}
+z_i\mathcal L_{\mathrm{RL}}^x(i).
\end{equation}
When $z_i=0$, prompt $x_i$ adopts the trace co-training loss in Eq.~\ref{eq:trace-cotraining-objective}.
When $z_i=1$, it leverages only the standard prompt-only RL loss $\mathcal{L}^{x}_{\mathrm{RL}}(i)$, without trace-conditioned likelihood evaluation or SFT supervision.
This retains trace-guided co-training while preserving a direct optimization path under the prompt-only condition used at inference time.

%% file: sections/5-experiment.tex
\section{Experiments}

\begin{table*}[t]
\centering
\small
\setlength{\tabcolsep}{4.5pt}
\renewcommand{\arraystretch}{1.08}
\caption{Main results on FTRL, BFCL, and ToolHop. AC denotes Answer Correctness. BFCL Avg averages Multi-Turn, Search, and Memory; the final Avg is the equal-weight mean of FTRL Solve-F1, BFCL Avg, and ToolHop AC.}
\resizebox{\textwidth}{!}{
\begin{tabular}{c|ccc|cccc|c|c}
\toprule
\multirow{2}{*}{\textbf{Methods}}
& \multicolumn{3}{c|}{\textbf{FTRL}}
& \multicolumn{4}{c|}{\textbf{BFCL}}
& \textbf{ToolHop} 
& \multirow{2}{*}{Avg} \\
\cmidrule(lr){2-4} \cmidrule(lr){5-8} \cmidrule(lr){9-9}
& Solve-R & Solve-P & Solve-F1
& Multi-Turn & Search & Memory & Avg
& AC & \\
\midrule
\rowcolor{gray!20}
\multicolumn{10}{c}{\textit{\textbf{Qwen3-1.7B}}} \\
\midrule
Vanilla  & 21.66 & 20.78 & 19.72 & 13.00 & 1.00  & 8.60  & 7.53  & 25.63 & 17.63 \\
SFT      & 22.36 & 17.89 & 17.66 & 13.38 & 2.50  & 8.60  & 8.16  & 23.32 & 16.38 \\
GRPO     & 21.57 & 17.37 & 18.13 & 11.88 & 3.50  & 9.25  & 8.21  & 25.13 & 17.16 \\
FTRL     & 22.78 & 22.06 & 21.06 & 12.62 & 6.00  & 11.18 & 9.93  & 26.43 & 19.14 \\
ToolRL   & 22.79 & 17.93 & 17.99 & 13.30 & 2.50  & 8.39  & 8.06  & 25.83 & 17.29 \\
CHORD    & 22.68 & 19.14 & 18.50 & 12.62 & 3.50  & 6.45  & 7.52  & 27.14 & 17.72 \\
OPSD & 25.24 & 20.60 & 21.15 & 12.62 & 3.00 & 9.46 & 8.36  & 25.53 & 18.35 \\
MatchTIR & 24.12 & 21.50 & 21.43 & 14.00 & 4.50  & 9.03  & 9.18  & 32.26 & 20.96 \\
SFT$\rightarrow$MatchTIR & 23.39 & 19.64 & 19.72 & 13.63 & 5.50 & 11.18 & 10.10 & 31.06 & 20.29 \\
\midrule
\rowcolor{lightgreen}
\textbf{Ours} & \textbf{28.33} & \textbf{26.94} & \textbf{22.93} & \textbf{14.88} & \textbf{7.00} & \textbf{13.63} & \textbf{11.84} & \textbf{34.47} & \textbf{23.08} \\
\midrule
\rowcolor{gray!20}
\multicolumn{10}{c}{\textit{\textbf{Qwen3-4B}}} \\
\midrule
Vanilla  & 32.14 & 31.66 & 28.60 & 23.50 & 12.00 & 16.34 & 17.28 & 31.86 & 25.91 \\
SFT      & 28.19 & 29.32 & 24.67 & 22.62 & 4.00  & 12.90 & 13.17 & 30.25 & 22.70 \\
GRPO     & 33.12 & 31.80 & 30.84 & 22.00 & 9.00  & 15.48 & 15.49 & 34.47 & 26.93 \\
FTRL     & 31.98 & 33.24 & 31.34 & 22.38 & 9.00  & 16.77 & 16.05 & 43.02 & 30.14 \\
ToolRL   & 32.59 & 28.98 & 28.28 & 23.00 & 11.00 & 17.42 & 17.14 & 34.27 & 26.56 \\
CHORD    & 32.69 & 29.84 & 29.88 & 22.00 & 12.50 & 15.48 & 16.66 & 37.69 & 28.08 \\
OPSD & 33.61 & 25.36 & 27.61 & 22.50 & 12.50 & 13.55 & 16.18 & 43.82 & 29.20 \\
MatchTIR & 36.10 & 30.60 & 31.60 & 24.38 & 17.50 & 14.62 & 18.83 & 47.54 & 32.66 \\
SFT$\rightarrow$MatchTIR & 37.17 & 30.40 & 32.03 & 23.62 & 17.50 & 13.76 & 18.29 & 47.34 & 32.55 \\
\midrule
\rowcolor{lightgreen}
\textbf{Ours} & \textbf{42.41} & \textbf{34.69} & \textbf{36.60} & \textbf{27.50} & \textbf{18.00} & \textbf{18.28} & \textbf{21.26} & \textbf{49.55} & \textbf{35.80}\\
\bottomrule
\end{tabular}
}
\label{tab:main_results}
\end{table*}

\subsection{Experiment Settings}

\paragraph{Models and datasets.}
We conduct experiments on Qwen3-1.7B and Qwen3-4B \citep{yang2025qwen3}. 
We use prompts from the released FTRL~\citep{ye2025ftrl} training split for model training and expert-trace construction.
For the main experiments, expert traces are primarily generated with Qwen3-235B, then verified and segmented as described in Appendix~\ref{app:training_data_construction}. During training, we use a batch size of 256 and a maximum of 10 interaction turns. For \method{}, we set the SFT scale $\beta_{\mathrm{SFT}}=0.1$. We use a prompt-only anchoring probability of $q=0.4$ for the main results. The objective, trace-component, and SFT-schedule ablations use $q=0$ to isolate the effects of trace co-training. Additional training and evaluation details are provided in Appendix~\ref{app:more_implementation_details}.

\paragraph{Baselines and metrics.}
We compare our method against several representative baselines, including the vanilla model, SFT, GRPO \citep{shao2024deepseekmath}, FTRL \citep{ye2025ftrl}, ToolRL \citep{qian2026toolrl}, CHORD \citep{zhang2026onpolicy}, OPSD \citep{penaloza2026privileged}, MatchTIR \citep{qu2026matchtir}, and SFT$\rightarrow$MatchTIR. 
SFT, CHORD, OPSD, and SFT$\rightarrow$MatchTIR share the same verified expert-trace pool as \method{}; the remaining trained baselines do not use full expert traces as conditioning context or SFT targets (Appendix~\ref{app:data-access}). 
We evaluate all methods on three tool-use benchmarks: FTRL \citep{ye2025ftrl}, BFCL \citep{patil2025bfcl}, and ToolHop \citep{ye2025toolhop}.  
We report solve rate (Solve-R), solve precision (Solve-P), and Solve F1 for FTRL, accuracy for BFCL, and Answer Correctness (AC) for ToolHop. 
Metric definitions are provided in Appendix~\ref{app:more_implementation_details}.

\subsection{Main Results}

Table~\ref{tab:main_results} reports the main results on FTRL, BFCL, and ToolHop. \method{} achieves the highest overall average score on both model sizes, improving over MatchTIR from 20.96 to 23.08 on Qwen3-1.7B and from 32.66 to 35.80 on Qwen3-4B. On FTRL, \method{} improves Solve-F1 over the best baseline for each model size: from 21.43 to 22.93 over MatchTIR on Qwen3-1.7B, and from 32.03 to 36.60 over SFT$\rightarrow$MatchTIR on Qwen3-4B. It also achieves the highest scores on all three reported BFCL subtasks and improves ToolHop AC from 32.26 to 34.47 and from 47.54 to 49.55 over MatchTIR on the two model sizes, respectively. \method{} also consistently outperforms OPSD, improving the overall average from 18.35 to 23.08 on Qwen3-1.7B and from 29.20 to 35.80 on Qwen3-4B.

These results support our main design choice. Compared with SFT and SFT$\rightarrow$MatchTIR, \method{} incorporates expert traces into the RL update and progressively relaxes direct imitation through component-aware SFT. Compared with prompt-only RL baselines such as FTRL and MatchTIR, \method{} provides additional guidance through expert-trace context and selective process supervision. The consistent gains show that keeping rollouts prompt-only while using traces only during optimization is effective for multi-turn tool-use learning. 

Additional experiments provide further evidence for these results. \method{} maintains its advantage over MatchTIR when both methods are trained for 48 updates on Qwen3-1.7B (Appendix~\ref{app:extended-training}). \method{} remains effective across different teacher models, with consistent gains over SFT$\rightarrow$MatchTIR (Appendix~\ref{app:teacher-quality}). Offline trace-construction and model-training costs are reported in Appendix~\ref{app:compute-cost}.

\begin{table}[!htbp]
\centering
\small
\caption{Ablation on training objectives using Qwen3-4B on FTRL and BFCL, with prompt-only anchoring disabled ($q=0$). Both combined objectives use the same component-aware SFT loss. BFCL Avg averages Multi-Turn, Search, and Memory accuracy.}
\begin{tabular}{c|ccc|c}
\toprule
& \multicolumn{3}{c|}{\textbf{FTRL}} & \textbf{BFCL} \\
\cmidrule(lr){2-4}\cmidrule(lr){5-5}
\textbf{Objective} & \textbf{Solve-R} & \textbf{Solve-P} & \textbf{Solve-F1} & \textbf{Avg} \\
\midrule
Vanilla & 32.14 & 31.66 & 28.60 & 17.28 \\
\midrule
$\mathcal{L}^{x}_{\mathrm{RL}}$ & 36.10 & 30.60 & 31.60 & 18.83 \\
$\mathcal{L}^{h}_{\mathrm{RL}}$ & 35.29 & 31.42 & 31.85 & 19.07 \\
$\mathcal{L}^{x}_{\mathrm{RL}}$+$\beta_{\mathrm{SFT}}\mathcal{L}_{\mathrm{SFT}}$ & 36.36 & 28.46 & 30.18 & 19.38 \\
$\mathcal{L}^{h}_{\mathrm{RL}}$+$\beta_{\mathrm{SFT}}\mathcal{L}_{\mathrm{SFT}}$ & \textbf{40.45} & \textbf{32.94} & \textbf{33.21} & \textbf{20.44} \\
\bottomrule
\end{tabular}
\label{tab:loss-ablation}
\end{table}

\subsection{Complementarity of Trace-Conditioned RL and SFT}
\label{sec training_objective}

To assess whether trace-conditioned RL and component-aware SFT provide complementary benefits, we compare prompt-only and trace-conditioned RL, each with and without component-aware SFT, on FTRL and BFCL with anchoring disabled ($q=0$; Table~\ref{tab:loss-ablation}).
On FTRL, trace-conditioning alone yields a modest improvement in Solve-F1 from 31.60 to 31.85, with higher Solve-P but lower Solve-R.
Adding component-aware SFT to prompt-only RL instead reduces Solve-F1 from 31.60 to 30.18 and Solve-P from 30.60 to 28.46.
In contrast, adding component-aware SFT to trace-conditioned RL increases Solve-F1 from 31.85 to 33.21, with improvements in both Solve-R and Solve-P.
Holding the SFT objective fixed, replacing prompt-only RL with trace-conditioned RL improves all three FTRL metrics, including a 3.03-point gain in Solve-F1.
On BFCL, adding component-aware SFT improves the average score by 0.55 points with prompt-only RL and by 1.37 points with trace-conditioned RL, with the joint objective achieving the highest average of 20.44.
These comparisons support the empirical complementarity of the two training signals across benchmarks: direct expert supervision is more beneficial when combined with trace-conditioned RL.

To examine whether trace conditioning remains beneficial in the main configuration ($q=0.4$), we compare \method{} with a variant that replaces trace-conditioned RL with prompt-only RL while retaining the same component-aware SFT and anchoring. Compared with this variant, \method{} improves FTRL Solve-F1 by 4.73 and BFCL Avg by 4.81. These gains support the value of using expert traces as privileged RL context beyond their role as SFT supervision in the main training configuration. To further assess the benefit of task-matched expert context within joint SFT--RL training, we conduct a controlled single-iteration comparison of different RL conditioning contexts. Details of both the full-training comparison and this controlled experiment are provided in Appendix~\ref{app:cross-benchmark-objectives}.

\begin{table}[!t]
\centering
\small
\setlength{\tabcolsep}{3.5pt}
\caption{Ablation of privileged RL context using Qwen3-4B on FTRL and BFCL ($q=0$). RE, TC, and TR denote reasoning, tool-calls, and tool responses. \xmark{} marks the removed component; SFT remains unchanged. BFCL Avg averages Multi-Turn, Search, and Memory accuracy.}
\begin{tabular}{ccc|ccc|c}
\toprule
\multicolumn{3}{c|}{} & \multicolumn{3}{c|}{\textbf{FTRL}} & \textbf{BFCL} \\
\cmidrule(lr){4-6}\cmidrule(lr){7-7}
\textbf{RE} & \textbf{TC} & \textbf{TR}
& Solve-R & Solve-P & Solve-F1 & Avg \\
\midrule
\xmark &  &  & 35.31 & 27.46 & 26.12 & 17.89 \\
 & \xmark &  & 39.19 & 28.52 & 31.11 & 16.31 \\
 &  & \xmark & 31.32 & 25.27 & 25.94 & 14.93 \\
\midrule
\multicolumn{3}{c|}{\textbf{All}} & \textbf{40.45} & \textbf{32.94} & \textbf{33.21} & \textbf{20.44} \\
\bottomrule
\end{tabular}
\label{tab:hint-ablation}
\end{table}

\subsection{The Role of Privileged Trace Components as RL Context}

Beyond the choice of training objectives, we examine which components of the expert trace provide useful context for RL policy optimization. 
We separately remove reasoning, tool-calls, and tool responses from the context used to compute the RL policy ratios, while keeping the SFT inputs, supervision masks, and loss coefficients unchanged (Table~\ref{tab:hint-ablation}). 
Removing reasoning or tool responses reduces Solve-F1 from 33.21 to 26.12 and 25.94, respectively, highlighting the value of both expert reasoning and environmental feedback as context for policy optimization. 
Removing tool-call context also lowers Solve-F1 to 31.11, indicating that expert action information remains useful even when tool-call SFT supervision is retained. 
Retaining all three components achieves the highest scores across all three FTRL metrics, supporting the use of complete expert traces as privileged context for RL updates. Detailed results are reported in Appendix~\ref{app:cross-benchmark-trace-components}.

\subsection{Choice of SFT Schedule}

We next examine how to schedule direct SFT supervision for reasoning and tool-calls, while keeping the trace-conditioned RL setup unchanged (Table~\ref{tab:sft_schedule}).
With tool-call loss-scale annealing fixed, segment-level annealing for reasoning achieves a Solve-F1 of 33.21, compared with 30.60 for fixed supervision and 29.86 for token-level annealing.
This suggests that gradually shortening the supervised reasoning prefixes at semantic boundaries is more effective than the compared alternatives.
For tool-calls, with segment-level reasoning annealing fixed, loss-scale annealing improves Solve-F1 by 0.39 points over a fixed weight and by 4.38 points over token-level annealing.
This pattern supports supervising complete tool-call spans throughout training while gradually relaxing the imitation constraint.
The combination achieves the highest scores across all three FTRL metrics among the compared schedules, supporting different annealing strategies for reasoning and tool-calls.
We provide additional schedule comparisons in Appendix~\ref{app:more_sft_schedule}.

\begin{table}[!htbp]
\centering
\small
\caption{Comparison of SFT schedules for reasoning and tool-calls using Qwen3-4B on FTRL with $q=0$. “Fix” denotes no annealing for the corresponding component, “Seg A” anneals reasoning supervision by semantic segment, “Token A” anneals supervision by token count, and “Scale A” anneals the tool-call loss weight.}
\begin{tabular}{cc|ccc}
\toprule
\textbf{Reasoning} & \textbf{Tool-Call} & \textbf{Solve-R} & \textbf{Solve-P} & \textbf{Solve-F1} \\
\midrule
Fix & Scale A & 37.90 & 28.40 & 30.60 \\
Seg A & Fix & 39.32 & 31.25 & 32.82 \\
Token A & Scale A & 34.77 & 28.08 & 29.86 \\
Seg A & Token A & 35.80 & 27.45 & 28.83 \\
\midrule
\textbf{Seg A} & \textbf{Scale A} & \textbf{40.45} & \textbf{32.94} & \textbf{33.21} \\
\bottomrule
\end{tabular}
\label{tab:sft_schedule}
\end{table}

\subsection{Effect of Prompt-only Anchoring}

\begin{figure}[!tbp]
    \centering
    \includegraphics[width=0.55\linewidth]{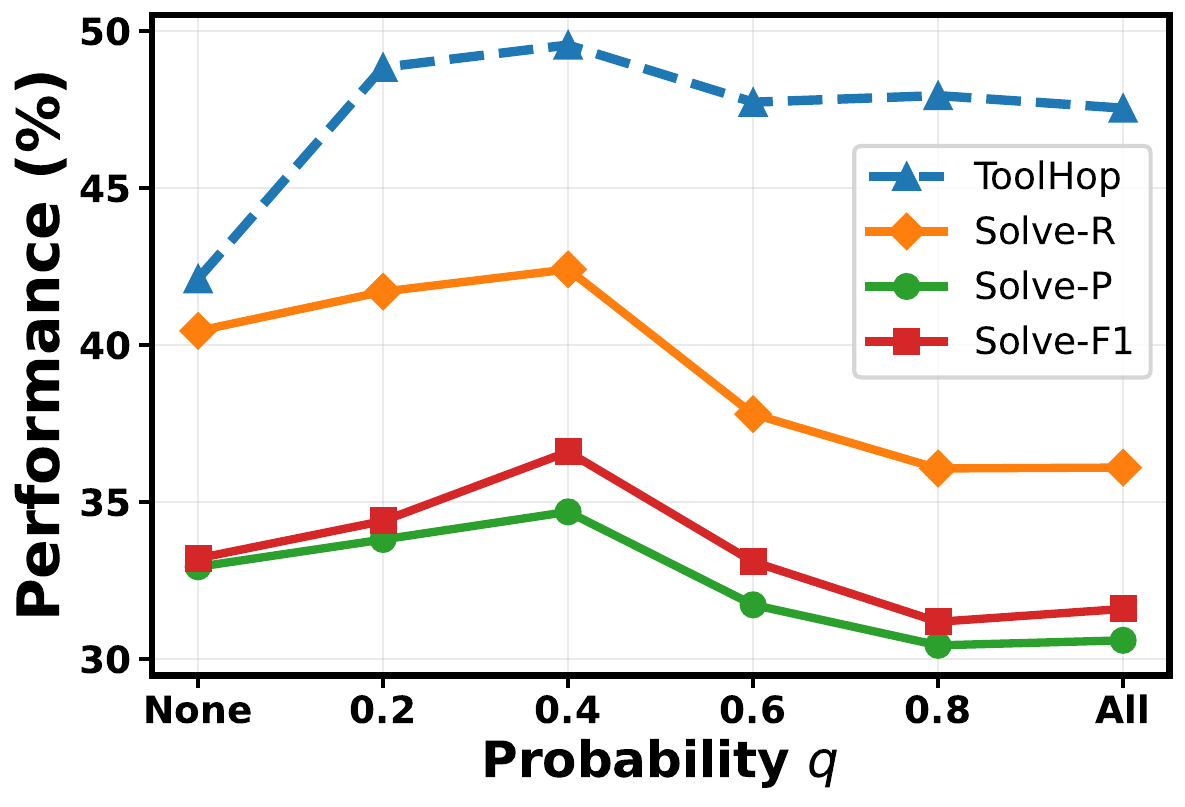}
    \caption{Effect of the prompt-only anchoring probability $q$ on Qwen3-4B. Curves show FTRL Solve-R, Solve-P, and Solve-F1, and ToolHop AC. ``None'' and ``All'' denote $q=0$ and $q=1$, respectively.}
    \label{fig:ratio}
\end{figure}

We vary the prompt-only anchoring probability $q$ to examine the balance between trace co-training and prompt-only RL (Figure~\ref{fig:ratio}). 
Without prompt-only anchoring, optimization relies entirely on trace co-training, which may encourage reliance on the training-only trace context. 
Introducing a moderate proportion of prompt-only RL improves ToolHop performance, suggesting that direct optimization under the test-time input condition helps the policy generalize to prompt-only use. 
However, when $q$ is large, fewer prompts receive trace-conditioned RL and component-aware SFT, and FTRL performance drops.

%% file: sections/6-conclusion.tex
\section{Conclusion}

In this work, we introduced \method{}, a privileged trace co-training framework for multi-turn tool-use agents.
\method{} keeps rollout generation prompt-only and uses expert traces during optimization through trace-conditioned RL and component-aware SFT.
Prompt-only anchoring adds a direct update signal under the inference-time input condition.
Experiments on FTRL, BFCL, and ToolHop show improvements over the SFT- and RL-based baselines, highlighting the value of expert traces for optimization without using them as rollout hints.

%% file: sections/7-appendix.tex
\section{More Implementation Details}
\label{app:more_implementation_details}

\subsection{Training Settings}
\label{app:training-settings}

Our implementation is based on the VeRL framework \citep{sheng2024hybridflow}. We use GRPO-style policy optimization with the rewards and advantages described in Appendix~\ref{app:training_reward}. For each training prompt, we sample $K=8$ prompt-only rollouts at temperature 1.0, with a maximum of 10 interaction turns per rollout. Policy-model training and evaluation use four NVIDIA H200 141GB GPUs. The table below summarizes the default configuration for the main \method{} experiments.

\begin{table}[htbp]
\centering
\small
\caption{Default training configuration for the main \method{} experiments. The temperature and rollout count refer to training-time sampling.}
\begin{tabular}{l c}
\toprule
\textbf{Hyperparameter} & \textbf{Value} \\
\midrule
\multicolumn{2}{l}{\textbf{Data and Token Budgets}} \\
Training batch size & 256 \\
Validation batch size & 256 \\
Maximum prompt length & 7000 \\
Maximum response length & 23000 \\
Maximum token length per GPU & 30000 \\
\midrule
\multicolumn{2}{l}{\textbf{Optimization}} \\
Learning rate & $1\times 10^{-6}$ \\
PPO mini-batch size & 32 \\
KL loss coefficient $\beta_{\mathrm{KL}}$ & 0.001 \\
SFT loss scale $\beta_{\mathrm{SFT}}$ & 0.1 \\
Reasoning loss scale & 1.0 \\
Tool-call start scale $\lambda_a^{\mathrm{start}}$ & 0.8 \\
Tool-call target end scale $\lambda_a^{\mathrm{end}}$ & 0.2 \\
Prompt-only anchoring probability $q$ & 0.4 \\
\midrule
\multicolumn{2}{l}{\textbf{Training Rollouts}} \\
Rollout framework & vLLM \\
Sampling temperature & 1.0 \\
Rollouts per prompt $K$ & 8 \\
Maximum interaction turns & 10 \\
GPU memory utilization & 0.7 \\
\midrule
\multicolumn{2}{l}{\textbf{Default Training Budget}} \\
Training updates & 24 \\
\bottomrule
\end{tabular}
\end{table}

We fix the overall SFT coefficient at $\beta_{\mathrm{SFT}}=0.1$ throughout training for \method{} and the objective-ablation variants that include component-aware SFT. The reasoning and tool-call losses are normalized separately.  The reasoning loss scale remains 1.0, while the supervised prefix within each assistant turn is annealed from $L_4$ to $L_0$ over five equal intervals of training progress. At $L_0$, reasoning tokens receive no direct SFT loss but remain in the context. Complete tool-call spans remain supervised, with their loss weight linearly annealed from 0.8 toward a target of 0.2. Prefix-level definitions are provided in Appendix~\ref{app:data-prefix-details}; additional SFT schedule comparisons are provided in Appendix~\ref{app:more_sft_schedule}.

For the main experiments, \(q=0.4\) was selected based on training-time performance. We compare $q\in\{0,0.2,0.4,0.6,0.8,1\}$ in Figure~\ref{fig:ratio}. The objective, privileged-context, and SFT-schedule ablations use $q=0$ to disable prompt-only anchoring. The extended-training experiment in Appendix~\ref{app:extended-training} increases the training budget from 24 to 48 updates.

\paragraph{Privileged-context construction.} 
We first format the task prompt using the model’s chat template. Expert reasoning, tool calls, and formatted tool responses are tokenized separately, and their token IDs are concatenated after the formatted prompt without reapplying the chat template or adding an instruction introducing the expert trace. For likelihood computation, the original prompt-only rollout is appended after this privileged context. The expert trace is supplied as a single prefix, rather than interleaved with individual rollout turns. The literal strings user and assistant in the expert tool-response formatting are ordinary text, not chat-template role delimiters.

\subsection{Baseline Implementation Details}
\label{app:baseline-implementation}

All baselines in Table~\ref{tab:main_results} start from the same Qwen3-1.7B or Qwen3-4B checkpoints. Table~\ref{tab:baseline-implementation} gives the actual training budgets and the key hyperparameters of each baseline. For the non-OPSD on-policy baselines, the common rollout temperature is 1.0, the PPO mini-batch size is 32, the reference-policy KL coefficient is 0.001, and the entropy coefficient is 0.001.

\begin{table}[htbp]
\centering
\small
\caption{Baseline training parameters for the main comparison. $B$ is the training batch size, $K$ is the number of rollouts per prompt, and $\eta$ is the policy learning rate. Updates refer to optimizer steps, not rollouts.}
\label{tab:baseline-implementation}
\begin{tabular}{@{}p{0.22\linewidth}p{0.12\linewidth}p{0.57\linewidth}@{}}
\toprule
\textbf{Baseline} & \textbf{Updates} & \textbf{Key training parameters} \\
\midrule
Vanilla & 0 & Unmodified Qwen3 checkpoint. \\
SFT & 24 & $B=256$, $\eta=10^{-6}$; full-trace cross-entropy. \\
GRPO & 24 & $B=256$, $K=8$, $\eta=10^{-6}$; final-answer exact-match reward. \\
FTRL & 24 & $B=256$, $K=8$, $\eta=10^{-6}$; Solve-F1 reward. \\
ToolRL & 24 & $B=256$, $K=8$, $\eta=10^{-6}$; tool-format and call-correctness reward. \\
CHORD & 24 & $B=256$, $K=8$, $\eta=10^{-6}$; fixed $\mu=0.1$ and token weight $\phi(p)=p(1-p)$ on expert SFT. \\
OPSD & 24 & $B=256$, $K=8$, $\eta=10^{-6}$, reverse-KL weight $\beta=0.1$. \\
MatchTIR & 24 & $B=256$, $K=8$, $\eta=10^{-6}$; tool-call matching and dual-level advantages ($\gamma=0.9$). \\
SFT$\rightarrow$MatchTIR & $24+24$ & Full-trace SFT then MatchTIR RL; $B=256$, $\eta=10^{-6}$, and $K=8$ in the RL stage. \\
\bottomrule
\end{tabular}
\end{table}

\subsection{Evaluation Settings}

We evaluate on the released FTRL test split \citep{ye2025ftrl}, the Multi-Turn, Search, and Memory subtasks of BFCL \citep{patil2025bfcl}, and ToolHop \citep{ye2025toolhop}. At evaluation time, the policy receives the task prompt and its ordinary interaction history, without privileged expert traces. Data-split separation and baseline access to expert traces are described in Appendix~\ref{app:data-access}.

\paragraph{Evaluation protocol.}
We perform one evaluation run per test instance on each benchmark. FTRL and ToolHop use greedy decoding (temperature 0) with thinking enabled and at most 10 model responses per instance; a response may contain multiple tool-calls. ToolHop is evaluated in the Free setting with a per-generation output limit of 4,096 tokens. BFCL uses a per-generation output cap of 4,096 tokens and follows the benchmark’s native interaction and step limits.

\paragraph{FTRL.}
For test instance $i$, let $n_i$ be the number of sub-questions, $p_i$ the number of tool invocations, and $c_i$ the number of successfully solved sub-questions. The per-instance solve precision $P_i$, solve rate $R_i$, and Solve-F1 $F_i$ are
\begin{align}
P_i=
\begin{cases}
c_i/p_i, & p_i>0,\\
1, & p_i=0,
\end{cases}
\qquad
R_i=\frac{c_i}{n_i},
\end{align}
\begin{align}
F_i=
\begin{cases}
\displaystyle\frac{2P_iR_i}{P_i+R_i}, & P_i+R_i>0,\\
0, & P_i+R_i=0.
\end{cases}
\end{align}
Solve-P measures tool-use precision, Solve-R measures task completeness, and Solve-F1 combines the two.

\paragraph{BFCL.}
We use the benchmark's accuracy-based evaluation for Multi-Turn, Search, and Memory. Unless otherwise specified, BFCL Avg is the arithmetic mean of these three subtask scores.

\paragraph{ToolHop.}
We report Answer Correctness (AC). For a test instance with reference answer $a_i$ and final response $o_i$, the per-instance score is 1 if $o_i$ contains $a_i$, and 0 otherwise:
\begin{align}
\mathrm{AC}_i=\mathbf{1}\!\left[\operatorname{contains}(o_i,a_i)\right].
\end{align}

\paragraph{Overall average.}
Unless otherwise specified, all reported scores are percentages. The final Avg in Table~\ref{tab:main_results} gives equal weight to the three benchmarks:
\begin{align}
\mathrm{Avg}
=
\frac{\mathrm{FTRL\ Solve\text{-}F1}
+\mathrm{BFCL\ Avg}
+\mathrm{ToolHop\ AC}}{3}.
\end{align}

\section{Training Reward and Advantage}
\label{app:training_reward}

We build on the matching-based rewards and dual-level advantage estimation of MatchTIR~\citep{qu2026matchtir}, with the implementation details specified below.
Both prompt-only RL and trace-conditioned RL use the same reward targets and advantage estimates.
For clarity, we fix a prompt $x_i$ and omit its index $i$ below.

\paragraph{Hard credit assignment.}
For each rollout, let $\mathcal P=\{p_1,\ldots,p_m\}$ denote its predicted tool-calls.
The reference tool-call list $\mathcal G=\{g_1,\ldots,g_n\}$ contains the fixed ground-truth calls associated with $x_i$ and supplies the matching targets for reward computation.
These targets are shared across the teacher variants in Appendix~\ref{app:teacher-quality} and are not replaced when the teacher-generated trace $h_i^*$ changes.
For each pair $(p_a,g_b)$, we compute a similarity score $S_{ab}$ based on tool-name match, argument-key overlap, and exact argument-value match.
Our implementation uses greedy one-to-one matching: candidate pairs are sorted by decreasing similarity, and a pair is accepted only if its score is positive and neither tool-call has already been matched.
Let $X_{ab}=1$ for an accepted pair and $0$ otherwise. The reward for $p_a$ is
\begin{align}
    r(p_a)=
    \begin{cases}
    S_{ab}, & \exists b,\ X_{ab}=1,\\
    -\lambda_{\mathrm{pen}}, & \text{otherwise},
    \end{cases}
    \qquad \lambda_{\mathrm{pen}}=0.
\end{align}

\paragraph{Turn-level and outcome rewards.}
For a tool-use turn $t$, we average the rewards of the predicted tool-calls in that turn:
\begin{align}
    r^{\mathrm{tool}}_t
    =\frac{1}{|\mathcal P_t|}\sum_{p\in\mathcal P_t}r(p),
\end{align}
with $r^{\mathrm{tool}}_t=0$ when $\mathcal P_t$ is empty.
For the final-answer turn, we use $r^{\mathrm{ans}}=\mathrm{F1}(\hat y,y)$ between the predicted answer $\hat y$ and reference answer $y$.
The per-turn reward $r_t$ is the tool-call reward for tool-use turns and the answer reward for the final turn.

\paragraph{Advantage estimation.}
Given $K$ rollouts for the same prompt, let $T_k$ be the number of turns in rollout $k$.
We compute its trajectory-level reward and normalize it within the rollout group:
\begin{equation}
    R_k=\sum_{t=1}^{T_k}r_{k,t},
    \qquad
    A^g_k=\frac{R_k-\operatorname{mean}_{k'=1,\ldots,K}(R_{k'})}
    {\operatorname{std}_{k'=1,\ldots,K}(R_{k'})+\epsilon_A}.
\end{equation}
For each turn, the discounted future reward is
\begin{equation}
    R^\ell_{k,t}=\sum_{u=t}^{T_k}\gamma^{u-t}r_{k,u},
    \qquad \gamma=0.9,\quad\epsilon_A=10^{-6}.
\end{equation}
Turn-level normalization uses only rollouts that reach turn $t$, indexed by $\mathcal K_t=\{k:T_k\ge t\}$:
\begin{equation}
    A^\ell_{k,t}=\frac{R^\ell_{k,t}-\mu_t}{\sigma_t+\epsilon_A},
    \qquad k\in\mathcal K_t.
\end{equation}
When $|\mathcal K_t|>1$, $\mu_t$ and $\sigma_t$ are the mean and sample standard deviation of $\{R^\ell_{k',t}:k'\in\mathcal K_t\}$.
Terminated rollouts are excluded from these statistics rather than zero-padded.
When $|\mathcal K_t|=1$, we set $\mu_t=0$ and $\sigma_t=1$; thus, the remaining rollout receives $A^\ell_{k,t}=R^\ell_{k,t}/(1+\epsilon_A)$.
For any model-generated token $n$ in turn $t$, the final advantage averages the two components:
\begin{equation}
    A_{k,n}=\frac{A^g_k+A^\ell_{k,t}}{2}.
\end{equation}
Tool-response tokens are generated by the environment and are excluded from the policy loss.

\section{Training Dataset Construction}
\label{app:training_data_construction}

We construct the training dataset from executable multi-turn tool-use tasks. Each
raw example contains a user question, a list of available tool schemas, the
corresponding executable tool implementations, and a reference answer. Since
\method{} requires process-level guidance rather than only final-answer labels, we
first generate expert tool-use trajectories with a teacher model, and
then convert these trajectories into segmented privileged traces for masked
process supervision.

\subsection{Data Sources and Access}
\label{app:data-access}

Expert-trace construction and model training use only released FTRL training prompts. FTRL evaluation uses its released test split; BFCL and ToolHop are used only for evaluation.
Among the trained baselines, GRPO, FTRL, ToolRL, and MatchTIR do not use full expert traces as policy-conditioning context or direct SFT targets. SFT, CHORD, OPSD, and SFT$\rightarrow$MatchTIR use the same verified expert-trace pool as \method{}. 
This distinction concerns full-trace conditioning and direct supervision; the fixed reference tool-calls used as reward targets are described in Appendix~\ref{app:training_reward}.

\subsection{Expert-Trace Construction}
\label{app:trace-construction}

\paragraph{Expert trajectory generation.}
For each raw question, we prompt a teacher model to solve the task under the same
tool-call interface used during training and inference. The teacher is
instructed to obtain all information through tool-calls and to output the final
answer only inside the \texttt{<answer></answer>} tag. After each generated tool-call, we execute the corresponding local tool implementation and append the tool
observation back to the context. This iterative process continues until the
teacher emits a final answer or reaches the maximum turn budget. We retain a
trajectory only when the final answer, or the last tool observation, verifies
against the reference answer.

We use the Qwen3-235B teacher generation as the main source of expert traces. We then remove invalid trajectories, including examples with invalid tool names, malformed
parsed turns, tool responses containing role text, or tool responses that
incorrectly contain tool-call tags. For training prompts without a verified Qwen3-235B trace, we additionally use xLAM-32B~\citep{zhang2025xlam} to improve trace coverage. Traces from both sources undergo the same execution-based verification and validity filtering before being included in the training set. The final training set contains 2,099 samples.

\paragraph{Hybrid segmentation of expert traces.}
To support component-aware SFT, we segment expert reasoning while retaining complete tool-call spans and tool observations in the trace. For each assistant turn, we extract the reasoning
inside the \texttt{<think>} block, normalize role noise and whitespace, split
the reasoning into paragraphs, and associate the turn with its structured tool-call and tool observation.

We apply a hybrid segmentation pipeline. Structurally simple turns are segmented
with deterministic rules, while harder turns are sent to an LLM annotator. If
both routes fail, we use a conservative fallback segmenter. Each segmented turn
is annotated with semantic chunks such as \texttt{task\_restate},
\texttt{tool\_scan}, \texttt{tool\_selection},
\texttt{required\_arg\_grounding}, \texttt{failure\_reflection}, and
\texttt{decision\_summary}. The segmenter also detects explicit tool-call
leakage, including literal JSON previews or phrases that introduce the exact
action. Such spans are marked as \texttt{explicit\_call\_preview} and are
excluded from eligible reasoning-supervision prefixes.

\subsection{Reasoning-Prefix Definitions and Supervision}
\label{app:data-prefix-details}

\paragraph{Eligible reasoning chunks.}
For each turn, we split the reasoning into ordered semantic chunks and retain the longest initial sequence of safe chunks ending immediately before the first explicit tool-call preview or other unsafe leakage span. Let this eligible sequence be $(C_1,\ldots,C_M)$. The levels are cumulative left prefixes, not five fixed semantic categories:
\begin{equation}
\begin{aligned}
L_0 &= \varnothing,\\
L_1 &= (C_1,\ldots,C_{\min(1,M)}),\\
L_2 &= (C_1,\ldots,C_{\min(2,M)}),\\
L_3 &= (C_1,\ldots,C_{\min(3,M)}),\\
L_4 &= (C_1,\ldots,C_M).
\end{aligned}
\label{eq:exact-prefix-levels}
\end{equation}
All tuples are empty when $M=0$. For $M\leq 3$, levels exceeding the available number of chunks repeat the longest eligible prefix. For $M>3$, $L_4$ includes all eligible chunks, rather than only the first four. The training schedule divides progress into five equal intervals and uses $L_4\rightarrow L_3\rightarrow L_2\rightarrow L_1\rightarrow L_0$.

\paragraph{Worked example and loss masking.}
Consider a span segmented into task restatement ($C_1$), tool selection ($C_2$), argument grounding ($C_3$), a pre-call decision ($C_4$), and a literal tool-call preview ($C_5$). In this example, the first four chunks are eligible and the preview is excluded from the eligible reasoning-supervision prefix, yielding $L_0=[]$, $L_1=[C_1]$, $L_2=[C_1,C_2]$, $L_3=[C_1,C_2,C_3]$, and $L_4=[C_1,C_2,C_3,C_4]$.

The selected prefix specifies which reasoning tokens receive direct SFT supervision in Eq.~\ref{eq:sft-mask}. The loss of tool-call tokens is normalized separately from the reasoning loss and then scaled by the scheduled weight $\lambda_a$. $L_0$ means no direct reasoning-token SFT loss. Rollout generation remains prompt-only as described in Section~\ref{sec:method-rollout-generation}.
\subsection{Prompt Templates}
\label{app:trace-prompt-templates}

\paragraph{Teacher prompt template.}
The prompt template used for expert trajectory generation is shown below. The
available tool schemas are rendered inside the \texttt{<tools></tools>} block,
and the teacher must serialize each function call inside
\texttt{<tool\_call></tool\_call>} \citep{qu2026matchtir}.

\begin{tcolorbox}[
  enhanced,
  breakable,
  width=\linewidth,
  colback=gray!8,
  colframe=black!70,
  coltitle=white,
  colbacktitle=black!75,
  title={Prompt Template for Expert Trajectory Generation},
  fonttitle=\bfseries,
  boxrule=0.8pt,
  arc=2pt,
  left=8pt,
  right=8pt,
  top=8pt,
  bottom=8pt,
  before skip=6pt,
  after skip=6pt
]
\raggedright
\small

\textbf{System Prompt:}

\# Tools

You may call one or more functions to assist with the user query.

You are provided with function signatures within
\texttt{<tools></tools>} XML tags:

\texttt{<tools>}

\{Tool List\}

\texttt{</tools>}

For each function call, return a json object with function name and arguments
within \texttt{<tool\_call></tool\_call>} XML tags:

\texttt{<tool\_call>}

\{\texttt{"name"}: \texttt{<function-name>},
\texttt{"arguments"}: \texttt{<args-json-object>}\}

\texttt{</tool\_call>}

\textbf{User Prompt:}

Please call given tools to answer the question. Please note that all your
information must be obtained by calling tools and not by answering the question
directly.

If the call fails, you need to try to correct it and continue until you arrive
at an answer.

Only output the final answer inside the \texttt{<answer></answer>} tag, without
any explanations or extra information.

Question: \{question\}

\textbf{Output:}
\end{tcolorbox}

\paragraph{LLM segmentation prompt.}
For difficult reasoning turns, we use an LLM annotator \footnote{We use ChatGPT-5.4 here.} to produce semantic
segments for reasoning-prefix supervision. The annotator is not asked to answer the
question; it only segments an existing teacher turn.

\begin{tcolorbox}[
  enhanced,
  breakable,
  width=\linewidth,
  colback=gray!8,
  colframe=black!70,
  coltitle=white,
  colbacktitle=black!75,
  title={Prompt Template for Trace Segmentation},
  fonttitle=\bfseries,
  boxrule=0.8pt,
  arc=2pt,
  left=8pt,
  right=8pt,
  top=8pt,
  bottom=8pt,
  before skip=6pt,
  after skip=6pt
]
\raggedright
\small

You are a data annotation engine for segmenting tool-use reasoning turns into
semantic chunks for autoregressive prefix-hint training.

Your job is NOT to answer the user question. Your job is to analyze one
assistant turn and split its reasoning into contiguous semantic chunks that can
be used as left-to-right prefix hints.

\textbf{Important training setting:}

- The downstream model is autoregressive.

- Only a LEFT PREFIX of the reasoning can be exposed as a hint.

- Therefore, segmentation must support prefix-level reveal.

- Chunks must preserve the original order.

- Chunks must be contiguous and non-overlapping.

\textbf{Main goal:}

Given the current turn's reasoning text, identify semantic stages such as task
understanding, tool scanning or comparison, tool selection, required argument
grounding, optional argument analysis, reflection after tool failure, fallback
planning, final decision summary before action, and explicit tool-call preview.

\textbf{Critical rule about leakage:}

If the reasoning contains an explicit preview of the exact tool call, JSON,
function name with all arguments, or a phrase that clearly introduces the
literal action, mark that span as \texttt{explicit\_call\_preview}. This span
should normally NOT be included in recommended prefix hint levels.

Return VALID JSON only. Do not include markdown. Do not paraphrase the original
reasoning text. Do not invent missing paragraphs. Use paragraph ids to define
chunks. If an intra-paragraph split is necessary, use
\texttt{split\_paragraph\_id} and an exact \texttt{cut\_before\_text} anchor.
Produce prefix levels \texttt{L0}, \texttt{L1}, \texttt{L2}, \texttt{L3}, and
\texttt{L4}, where \texttt{L0} reveals no hint and \texttt{L4} is the strongest
safe prefix before leakage.

\textbf{Allowed chunk types:}

\texttt{task\_restate}, \texttt{tool\_scan}, \texttt{tool\_selection},
\texttt{required\_arg\_grounding}, \texttt{optional\_arg\_analysis},
\texttt{irrelevant\_tool\_rejection}, \texttt{failure\_reflection},
\texttt{fallback\_plan}, \texttt{decision\_summary},
\texttt{explicit\_call\_preview}, \texttt{final\_answer\_grounding},
\texttt{other}.

\textbf{Allowed turn patterns:}

\texttt{initial\_attempt}, \texttt{compare\_then\_call},
\texttt{retry\_after\_failed\_call}, \texttt{post\_observation\_followup},
\texttt{final\_answer\_turn}, \texttt{other}.

\textbf{Return JSON following this schema:}

\{
  "turn\_pattern": "...",
  "cleaning\_actions": [\{"type": "...", "detail": "..."\}],
  "chunks": [
    \{
      "chunk\_id": "C1",
      "chunk\_type": "...",
      "paragraph\_ids": [0, 1],
      "split\_paragraph\_id": null,
      "cut\_before\_text": null,
      "summary": "...",
      "keep\_for\_prefix": true,
      "hint\_strength": "weak|medium|strong"
    \}
  ],
  "dangerous\_spans": [
    \{
      "type": "explicit\_call\_preview",
      "paragraph\_id": 4,
      "cut\_before\_text": "So, the tool call would be:"
    \}
  ],
  "recommended\_prefix\_levels": [
    \{"level": "L0", "visible\_chunks": []\},
    \{"level": "L1", "visible\_chunks": ["C1"]\},
    \{"level": "L2", "visible\_chunks": ["C1", "C2"]\},
    \{"level": "L3", "visible\_chunks": ["C1", "C2", "C3"]\},
    \{"level": "L4", "visible\_chunks": ["C1", "C2", "C3", "C4"]\}
  ],
  "quality\_flags": ["clean\_structure"]
\}
\end{tcolorbox}

\section{Training Algorithm}
\label{app:algorithm}

Algorithm~\ref{alg:trace} gives the training procedure of \method{}.
Rollouts are generated without expert-trace conditioning; rewards and advantages are computed as described in Appendix~\ref{app:training_reward}.
Each prompt selects either trace co-training or prompt-only RL, with one gate shared by its $K$ rollouts.
Component-aware SFT follows the masks and annealing rules in Section~\ref{sec:method-component-sft}, while retaining the full teacher-forced context.
The per-prompt SFT contributions are scaled so that their batch average recovers the separately normalized loss in Eq.~\ref{eq:sft-loss} over trace-branch prompts; a component with no selected tokens contributes zero.

\begin{algorithm}[H]
\caption{\method{} Training}
\label{alg:trace-balanced}
\label{alg:trace}
\small
\begin{algorithmic}[1]
\Require Policy $\pi_\theta$, reference policy $\pi_{\mathrm{ref}}$, training data $\{(x_i,h_i^*)\}$, rollout count $K$, anchoring probability $q$, SFT weight $\beta_{\mathrm{SFT}}$.
\For{each training update $s$}
    \State Freeze $\pi_{\mathrm{old}}\leftarrow\pi_\theta$ and sample a batch $\mathcal B$.
    \State Sample $\tau_{i,k}\sim\pi_{\mathrm{old}}(\cdot\mid x_i)$ for each $i\in\mathcal B$ and $k=1,\ldots,K$.
    \State Compute rollout rewards and token-level advantages $A_{i,k,n}$ (Appendix~\ref{app:training_reward}).
    \State Sample one gate $z_i\sim\operatorname{Bernoulli}(q)$ for each prompt.
    \State Update reasoning-prefix masks and tool-call weight $\lambda_a(s)$ (Section~\ref{sec:method-component-sft}).
    \State Compute SFT contributions $\mathcal L_{\mathrm{SFT}}(i;s)$ for $z_i=0$, using separate component normalization (Eq.~\ref{eq:sft-loss}).
    \For{each prompt $i\in\mathcal B$}
        \If{$z_i=0$}
            \State Set the privileged context $c_i^h=[x_i;h_i^*]$.
            \State $\displaystyle r^h_{i,k,n}\leftarrow\frac{\pi_\theta(\tau_{i,k,n}\mid c_i^h,\tau_{i,k,<n})}{\pi_{\mathrm{old}}(\tau_{i,k,n}\mid c_i^h,\tau_{i,k,<n})}$, $\quad n\in\mathcal M_{i,k}$.
            \State Compute $\mathcal L^h_{\mathrm{RL}}(i)$ from $r^h_{i,k,n}$ and $A_{i,k,n}$, with clipping and KL regularization (Eq.~\ref{eq:trace-rl-loss}).
            \State $\mathcal L_i^{\mathrm{trace}}\leftarrow\mathcal L^h_{\mathrm{RL}}(i)+\beta_{\mathrm{SFT}}\mathcal L_{\mathrm{SFT}}(i;s)$.
            \State $\mathcal L_i\leftarrow\mathcal L_i^{\mathrm{trace}}$.
        \Else
            \State Compute $\mathcal L^x_{\mathrm{RL}}(i)$ using prompt-only context and the same rollout advantages.
            \State $\mathcal L_i\leftarrow\mathcal L^x_{\mathrm{RL}}(i)$ \Comment{No SFT loss}
        \EndIf
    \EndFor
    \State Minimize $\displaystyle\mathcal L=\frac{1}{|\mathcal B|}\sum_{i\in\mathcal B}\mathcal L_i$ with an optimizer step on $\theta$.
\EndFor
\end{algorithmic}
\end{algorithm}

\section{A Conceptual Latent-Trace View of \method{}}
\label{app:latent-trace-conceptual}

We define an auxiliary latent-trace model to provide a conceptual interpretation of the two trace-based training signals in \method{}: learning from expert traces and using them as context for updating self-generated rollouts.

\paragraph{Auxiliary latent-trace model.}
For a fixed prompt $x$, let $p_\theta(h\mid x)$ be a normalized distribution over discrete process traces. Let $P_\theta(\tau\mid x,h)$ denote the trajectory distribution induced by $\pi_\theta$ conditioned on $[x;h]$ and an environment whose transition probabilities do not depend on $\theta$. We assume a finite interaction horizon and include the generation and termination rules in these distributions. We define the auxiliary mixture distribution
\begin{equation}
\widetilde P_\theta(\tau\mid x)
:=\sum_h p_\theta(h\mid x)P_\theta(\tau\mid x,h).
\label{eq:conceptual-latent-joint}
\end{equation}
The mixture $\widetilde P_\theta$ is introduced for this conceptual analysis; it is not assumed to equal the deployed prompt-only policy. \method{} generates rollouts directly from the prompt-only policy, without first sampling an auxiliary trace.

\paragraph{An expert-trace lower bound.}
For an expert trace $h^*$ with positive joint probability for the rollout under consideration, retaining its term in the marginal gives
\begin{equation}
\begin{aligned}
\log\widetilde P_\theta(\tau\mid x)
\geq\;&\underbrace{\log p_\theta(h^*\mid x)}_{\text{trace likelihood}}
&+\underbrace{\log P_\theta(\tau\mid x,h^*)}_{\text{trace-conditioned rollout likelihood}}.
\end{aligned}
\label{eq:conceptual-latent-bound}
\end{equation}
The two terms highlight complementary roles for an expert trace: learning the expert process itself and using that process as context for evaluating a sampled rollout. This provides a conceptual motivation for combining direct process supervision with trace-conditioned policy updates; the connections to the implemented objectives are detailed below.

\paragraph{Connection to component-aware SFT.}
The trace-likelihood term motivates learning the expert process itself. In \method{}, component-aware SFT serves this role by supervising selected reasoning prefixes and complete tool-call spans (Eq.~\ref{eq:sft-loss}). Tool observations and unsupervised reasoning tokens remain in the teacher-forced context. Selective masks, separate component normalization, and an annealed tool-call weight define a structured supervision objective for learning selected components of the expert process.

\paragraph{Connection to trace-conditioned RL.}
The conditional likelihood term motivates using the expert process as context for evaluating self-generated behavior. For a fixed rollout, the parameter-dependent factors in $P_\theta(\tau\mid x,h^*)$ are the policy-generated token probabilities, including any policy-controlled termination decisions; environment-transition factors contribute no parameter gradient. \method{} uses these token probabilities in its clipped policy ratios, retaining the advantages of the original prompt-only rollouts (Eq.~\ref{eq:trace-rl-loss}). 

\paragraph{Interpretation of joint training.}
This view motivates two related uses of the same expert trace: direct supervision of its model-generated components and privileged conditioning of reward-based updates on autonomous rollouts. \method{} couples these signals through shared policy parameters, and Table~\ref{tab:loss-ablation} evaluates their empirical complementarity. 

\newcommand{\turnsep}{\par\medskip\noindent\rule{\linewidth}{0.4pt}\medskip}
\newcommand{\bad}[1]{\textcolor{red}{#1}}
\newcommand{\good}[1]{\textcolor{green!50!black}{#1}}
\input{sections/8-additional-results}

\begingroup
\setlength{\parskip}{3pt}
\lstset{aboveskip=3pt,belowskip=3pt}
\section{Case Study}
\label{app:case_study}

We provide a case study to compare the multi-turn tool-use behavior of the vanilla model, MatchTIR, and \method{}.
For readability, we truncate the reasoning text but keep the full tool-calls, tool responses, and final answers.
Incorrect or less accurate parts are highlighted in red.

\begin{table}[!h]
\centering
\small
\caption{Case study comparison on the same prompt. Scores are reported on a 0–1 scale.}
\begin{tabular}{lccc}
\toprule
\textbf{Model}  & \textbf{Solve-P} & \textbf{Solve-R} & \textbf{Solve-F1} \\
\midrule
Vanilla Qwen3-4B  & 0.29 & 0.33 & 0.31 \\
MatchTIR  & \textbf{1.00} & 0.67 & 0.80 \\
\method{}  & \textbf{1.00} & \textbf{1.00} & \textbf{1.00} \\
\bottomrule
\end{tabular}

\label{tab:case_study}
\end{table}

\begin{tcolorbox}[
  enhanced,
  breakable,
  lines before break=6,
  width=\linewidth,
  colback=gray!8,
  colframe=black!70,
  coltitle=white,
  colbacktitle=black!75,
  title={Case Study Prompt},
  fonttitle=\bfseries,
  boxrule=0.8pt,
  arc=2pt,
  left=8pt,
  right=8pt,
  top=8pt,
  bottom=8pt,
  before skip=6pt,
  after skip=6pt
]
\raggedright
\small

\textbf{Question.}\par\nopagebreak[4]
What is the quotient of the area of the reserve located near the tribe whose ceremonial dance has been declared an intangible cultural heritage in hectares, and the birth year of the scientist who discovered oxygen?

\medskip
\textbf{Reference Answer (not provide).}\par\nopagebreak[4]
The correct reserve is Serengeti National Park, whose area is 1,500,000 hectares. The scientist who discovered oxygen is Joseph Priestley, whose birth year is 1733. Therefore, the quotient is:
\[
\frac{1,\!500,\!000}{1733} \approx 865.55.
\]

\textbf{Answer:} $\boxed{865.55}$

\end{tcolorbox}

\begin{tcolorbox}[
  enhanced,
  breakable,
  lines before break=6,
  width=\linewidth,
  colback=gray!8,
  colframe=black!70,
  coltitle=white,
  colbacktitle=black!75,
  title={Vanilla Qwen3-4B},
  fonttitle=\bfseries,
  boxrule=0.8pt,
  arc=2pt,
  left=8pt,
  right=8pt,
  top=8pt,
  bottom=8pt,
  before skip=6pt,
  after skip=6pt
]
\raggedright
\small

\textbf{Turn 1: Identify the tribe}\par\nopagebreak[4]
\textbf{Assistant Reasoning.}\par\nopagebreak[4]
\begin{lstlisting}[basicstyle=\ttfamily\small,breaklines=true,escapeinside={(*@}{@*)}]
Okay, let's tackle this question step by step. The user is asking for the quotient of two values: the area of a reserve near a tribe whose ceremonial dance is an intangible cultural heritage, in hectares, divided by the birth year of the scientist who discovered oxygen. ...
\end{lstlisting}

\textbf{Assistant Tool-Call.}\par\nopagebreak[4]
\begin{lstlisting}[basicstyle=\ttfamily\small,breaklines=true,escapeinside={(*@}{@*)}]
{"name": "cultural_heritage_locator", "arguments": {"heritage_type": "dance", "include_unesco": true}}
\end{lstlisting}

\textbf{Tool Response.}\par\nopagebreak[4]
\begin{lstlisting}[basicstyle=\ttfamily\small,breaklines=true,escapeinside={(*@}{@*)}]
The Maasai tribe is known for their traditional dance, which is a significant cultural heritage.
\end{lstlisting}

\turnsep

\textbf{Turn 2: Search for a nearby reserve}\par\nopagebreak[4]
\textbf{Assistant Reasoning.}\par\nopagebreak[4]
\begin{lstlisting}[basicstyle=\ttfamily\small,breaklines=true,escapeinside={(*@}{@*)}]
Okay, so the user wants the quotient of the area of a reserve near the Maasai tribe's location divided by the birth year of the oxygen discoverer. ...
\end{lstlisting}

\textbf{Assistant Tool-Call.}\par\nopagebreak[4]
\begin{lstlisting}[basicstyle=\ttfamily\small,breaklines=true,escapeinside={(*@}{@*)}]
{"name": "nearby_reserve_locator", "arguments": {"cultural_location": "Maasai", "region": "Tanzania"}}
\end{lstlisting}

\textbf{Tool Response.}\par\nopagebreak[4]
\begin{lstlisting}[basicstyle=\ttfamily\small,breaklines=true,escapeinside={(*@}{@*)}]
(*@\bad{No known reserves found near the specified location.}@*)
\end{lstlisting}

\turnsep

\textbf{Turn 3: Retry with Kenya}\par\nopagebreak[4]
\textbf{Assistant Reasoning.}\par\nopagebreak[4]
\begin{lstlisting}[basicstyle=\ttfamily\small,breaklines=true,escapeinside={(*@}{@*)}]
Okay, the user is asking for the quotient of the area of a reserve near the Maasai tribe's location and the birth year of the oxygen discoverer. But when I tried to find reserves near the Maasai in Tanzania, there were none. ...
\end{lstlisting}

\textbf{Assistant Tool-Call.}\par\nopagebreak[4]
\begin{lstlisting}[basicstyle=\ttfamily\small,breaklines=true,escapeinside={(*@}{@*)}]
{"name": "nearby_reserve_locator", "arguments": {"cultural_location": "Maasai", "region": "Kenya", "reserve_type": "cultural"}}
\end{lstlisting}

\textbf{Tool Response.}\par\nopagebreak[4]
\begin{lstlisting}[basicstyle=\ttfamily\small,breaklines=true,escapeinside={(*@}{@*)}]
(*@\bad{No known reserves found near the specified location.}@*)
\end{lstlisting}

\turnsep

\textbf{Turn 4: Retry with Maasai Mara}\par\nopagebreak[4]
\textbf{Assistant Reasoning.}\par\nopagebreak[4]
\begin{lstlisting}[basicstyle=\ttfamily\small,breaklines=true,escapeinside={(*@}{@*)}]
Okay, the user is asking for the quotient of the area of a reserve near the Maasai tribe's location and the birth year of the oxygen discoverer. But both attempts to find reserves near the Maasai in Tanzania and Kenya didn't return any results. ...
\end{lstlisting}

\textbf{Assistant Tool-Call.}\par\nopagebreak[4]
\begin{lstlisting}[basicstyle=\ttfamily\small,breaklines=true,escapeinside={(*@}{@*)}]
{"name": "nearby_reserve_locator", "arguments": {"cultural_location": "Maasai Mara", "region": "Kenya", "reserve_type": "cultural"}}
\end{lstlisting}

\textbf{Tool Response.}\par\nopagebreak[4]
\begin{lstlisting}[basicstyle=\ttfamily\small,breaklines=true,escapeinside={(*@}{@*)}]
(*@\bad{No known reserves found near the specified location.}@*)
\end{lstlisting}

\turnsep

\textbf{Turn 5: Look up birth year}\par\nopagebreak[4]
\textbf{Assistant Reasoning.}\par\nopagebreak[4]
\begin{lstlisting}[basicstyle=\ttfamily\small,breaklines=true,escapeinside={(*@}{@*)}]
Okay, the user is asking for the quotient of the area of a reserve near the Maasai tribe's location and the birth year of the oxygen discoverer. But all attempts to find reserves near the Maasai have failed. ...
\end{lstlisting}

\textbf{Assistant Tool-Call.}\par\nopagebreak[4]
\begin{lstlisting}[basicstyle=\ttfamily\small,breaklines=true,escapeinside={(*@}{@*)}]
{"name": "historical_fact_lookup", "arguments": {"person_name": "Joseph Priestley", "fact_type": "birth_year"}}
\end{lstlisting}

\textbf{Tool Response.}\par\nopagebreak[4]
\begin{lstlisting}[basicstyle=\ttfamily\small,breaklines=true,escapeinside={(*@}{@*)}]
{"result": "1733"}
\end{lstlisting}

\turnsep

\textbf{Turn 6: Use an incorrect reserve}\par\nopagebreak[4]
\textbf{Assistant Reasoning.}\par\nopagebreak[4]
\begin{lstlisting}[basicstyle=\ttfamily\small,breaklines=true,escapeinside={(*@}{@*)}]
Okay, let's see. The user is asking for the quotient of the area of a reserve near the Maasai tribe's location divided by the birth year of the oxygen discoverer. ...
\end{lstlisting}

\textbf{Assistant Tool-Call.}\par\nopagebreak[4]
\begin{lstlisting}[basicstyle=\ttfamily\small,breaklines=true,escapeinside={(*@}{@*)}]
{"name": "area_calculator", "arguments": {"location_name": "(*@\bad{Maasai Mara National Reserve}@*)", "unit": "hectares"}}
\end{lstlisting}

\textbf{Tool Response.}\par\nopagebreak[4]
\begin{lstlisting}[basicstyle=\ttfamily\small,breaklines=true,escapeinside={(*@}{@*)}]
{"location_name": "(*@\bad{Maasai Mara National Reserve}@*)", "area": (*@\bad{100000}@*), "unit": "hectares", "include_water_bodies": true, "precision_level": 3, "data_source": "satellite"}
\end{lstlisting}

\turnsep

\textbf{Turn 7: Final calculation}\par\nopagebreak[4]
\textbf{Assistant Reasoning.}\par\nopagebreak[4]
\begin{lstlisting}[basicstyle=\ttfamily\small,breaklines=true,escapeinside={(*@}{@*)}]
Okay, let's see. The user wants the quotient of the area of a reserve near the Maasai tribe's location divided by the birth year of the oxygen discoverer. ...
\end{lstlisting}

\textbf{Assistant Tool-Call.}\par\nopagebreak[4]
\begin{lstlisting}[basicstyle=\ttfamily\small,breaklines=true,escapeinside={(*@}{@*)}]
{"name": "advanced_calculator", "arguments": {"operation": "division", "operand1": (*@\bad{100000}@*), "operand2": 1733}}
\end{lstlisting}

\textbf{Tool Response.}\par\nopagebreak[4]
\begin{lstlisting}[basicstyle=\ttfamily\small,breaklines=true,escapeinside={(*@}{@*)}]
{"result": "(*@\bad{57.70}@*)"}
\end{lstlisting}

\textbf{Assistant Reasoning.}\par\nopagebreak[4]
\begin{lstlisting}[basicstyle=\ttfamily\small,breaklines=true,escapeinside={(*@}{@*)}]
Okay, let's wrap this up. The user asked for the quotient of the area of a reserve near the Maasai tribe's location (in hectares) divided by the birth year of the oxygen discoverer. ...
\end{lstlisting}

\textbf{Assistant Final Answer.}\par\nopagebreak[4]
\begin{lstlisting}[basicstyle=\ttfamily\small,breaklines=true,escapeinside={(*@}{@*)}]
The quotient of the area of the (*@\bad{Maasai Mara National Reserve}@*) ((*@\bad{100,000 hectares}@*)) divided by the birth year of Joseph Priestley (1733) is **(*@\bad{57.70}@*)**. 

**Answer:** (*@\bad{57.70}@*)
\end{lstlisting}

\turnsep

\textbf{Error Summary.}\par\nopagebreak[4]
\begin{lstlisting}[basicstyle=\ttfamily\small,breaklines=true,escapeinside={(*@}{@*)}]
(*@\bad{The vanilla model is derailed by failed reserve-search attempts and then switches to the wrong reserve, Maasai Mara National Reserve.}@*)
(*@\bad{As a result, it uses the wrong area, 100,000 hectares, instead of the correct Serengeti National Park area of 1,500,000 hectares.}@*)
(*@\bad{The birth year and calculator call are correct, but the final quotient is wrong because the reserve and area are incorrect.}@*)
\end{lstlisting}

\end{tcolorbox}

\begin{tcolorbox}[
  enhanced,
  breakable,
  lines before break=6,
  width=\linewidth,
  colback=gray!8,
  colframe=black!70,
  coltitle=white,
  colbacktitle=black!75,
  title={MatchTIR},
  fonttitle=\bfseries,
  boxrule=0.8pt,
  arc=2pt,
  left=8pt,
  right=8pt,
  top=8pt,
  bottom=8pt,
  before skip=6pt,
  after skip=6pt
]
\raggedright
\small

\textbf{Turn 1: Identify the tribe}\par\nopagebreak[4]
\textbf{Assistant Reasoning.}\par\nopagebreak[4]
\begin{lstlisting}[basicstyle=\ttfamily\small,breaklines=true,escapeinside={(*@}{@*)}]
Okay, let's tackle this question step by step. The user is asking for the quotient of two values: the area of a specific reserve near a tribe with a ceremonial dance declared as intangible cultural heritage, in hectares, divided by the birth year of the scientist who discovered oxygen. ...
\end{lstlisting}

\textbf{Assistant Tool-Call.}\par\nopagebreak[4]
\begin{lstlisting}[basicstyle=\ttfamily\small,breaklines=true,escapeinside={(*@}{@*)}]
{"name": "cultural_heritage_locator", "arguments": {"heritage_type": "dance", "include_unesco": true}}
\end{lstlisting}

\textbf{Tool Response.}\par\nopagebreak[4]
\begin{lstlisting}[basicstyle=\ttfamily\small,breaklines=true,escapeinside={(*@}{@*)}]
The Maasai tribe is known for their traditional dance, which is a significant cultural heritage.
\end{lstlisting}

\turnsep

\textbf{Turn 2: Locate the reserve}\par\nopagebreak[4]
\textbf{Assistant Reasoning.}\par\nopagebreak[4]
\begin{lstlisting}[basicstyle=\ttfamily\small,breaklines=true,escapeinside={(*@}{@*)}]
Okay, so the user wants the quotient of the area of a reserve near the Maasai tribe's ceremonial dance (which is intangible cultural heritage) in hectares, divided by the birth year of the scientist who discovered oxygen. ...
\end{lstlisting}

\textbf{Assistant Tool-Call.}\par\nopagebreak[4]
\begin{lstlisting}[basicstyle=\ttfamily\small,breaklines=true,escapeinside={(*@}{@*)}]
{"name": "nearby_reserve_locator", "arguments": {"cultural_location": "Maasai tribe", "radius": 100, "region": "Kenya"}}
\end{lstlisting}

\textbf{Tool Response.}\par\nopagebreak[4]
\begin{lstlisting}[basicstyle=\ttfamily\small,breaklines=true,escapeinside={(*@}{@*)}]
Serengeti National Park
\end{lstlisting}

\turnsep

\textbf{Turn 3: Get reserve area}\par\nopagebreak[4]
\textbf{Assistant Reasoning.}\par\nopagebreak[4]
\begin{lstlisting}[basicstyle=\ttfamily\small,breaklines=true,escapeinside={(*@}{@*)}]
Okay, so the user wants the quotient of the area of the Serengeti National Park (near the Maasai tribe) in hectares divided by the birth year of the scientist who discovered oxygen. ...
\end{lstlisting}

\textbf{Assistant Tool-Call.}\par\nopagebreak[4]
\begin{lstlisting}[basicstyle=\ttfamily\small,breaklines=true,escapeinside={(*@}{@*)}]
{"name": "area_calculator", "arguments": {"location_name": "Serengeti National Park", "unit": "hectares"}}
\end{lstlisting}

\textbf{Tool Response.}\par\nopagebreak[4]
\begin{lstlisting}[basicstyle=\ttfamily\small,breaklines=true,escapeinside={(*@}{@*)}]
{"location_name": "Serengeti National Park", "area": 1500000, "unit": "hectares", "include_water_bodies": true, "precision_level": 3, "data_source": "satellite"}
\end{lstlisting}

\turnsep

\textbf{Turn 4: Get birth year}\par\nopagebreak[4]
\textbf{Assistant Reasoning.}\par\nopagebreak[4]
\begin{lstlisting}[basicstyle=\ttfamily\small,breaklines=true,escapeinside={(*@}{@*)}]
Okay, so the area of Serengeti National Park is 1,500,000 hectares. Now I need to find the birth year of the scientist who discovered oxygen. From what I remember, the discovery of oxygen is often attributed to Joseph Priestley, but I should verify this using the historical_fact_lookup tool. ...
(*@\bad{Missing step: the model does not call element\_discovery\_info to verify Joseph Priestley is the oxygen discoverer.}@*)
\end{lstlisting}

\textbf{Assistant Tool-Call.}\par\nopagebreak[4]
\begin{lstlisting}[basicstyle=\ttfamily\small,breaklines=true,escapeinside={(*@}{@*)}]
{"name": "historical_fact_lookup", "arguments": {"person_name": "Joseph Priestley", "fact_type": "birth_year"}}
\end{lstlisting}

\textbf{Tool Response.}\par\nopagebreak[4]
\begin{lstlisting}[basicstyle=\ttfamily\small,breaklines=true,escapeinside={(*@}{@*)}]
{"result": "1733"}
\end{lstlisting}

\turnsep

\textbf{Turn 5: Final answer}\par\nopagebreak[4]
\textbf{Assistant Reasoning.}\par\nopagebreak[4]
\begin{lstlisting}[basicstyle=\ttfamily\small,breaklines=true,escapeinside={(*@}{@*)}]
Okay, so the area of Serengeti National Park is 1,500,000 hectares, and the birth year of Joseph Priestley is 1733. The user wants the quotient of these two values.

Wait, the quotient is area divided by birth year. So that's 1,500,000 / 1733. Let me calculate that. ...
(*@\bad{Missing step: the model does not call advanced\_calculator for the final division and instead computes an approximate value manually.}@*)
\end{lstlisting}

\textbf{Assistant Final Answer.}\par\nopagebreak[4]
\begin{lstlisting}[basicstyle=\ttfamily\small,breaklines=true,escapeinside={(*@}{@*)}]
The quotient is calculated by dividing the area of Serengeti National Park (1,500,000 hectares) by the birth year of Joseph Priestley (1733). 

$$
\frac{1,500,000}{1733} \approx (*@\bad{865.5}@*)
$$

**Final Answer:** (*@\bad{865.5}@*)
\end{lstlisting}

\turnsep

\textbf{Error Summary.}\par\nopagebreak[4]
\begin{lstlisting}[basicstyle=\ttfamily\small,breaklines=true,escapeinside={(*@}{@*)}]
(*@\bad{Although MatchTIR follows the main correct path, it misses two expected tool-use steps: verifying the oxygen discoverer with element\_discovery\_info and computing the final quotient with advanced\_calculator. Therefore, its answer is only approximate and its tool-use trajectory is incomplete.}@*)
\end{lstlisting}

\end{tcolorbox}

\begin{tcolorbox}[
  enhanced,
  breakable,
  lines before break=6,
  width=\linewidth,
  colback=gray!8,
  colframe=black!70,
  coltitle=white,
  colbacktitle=black!75,
  title={\method{} Output},
  fonttitle=\bfseries,
  boxrule=0.8pt,
  arc=2pt,
  left=8pt,
  right=8pt,
  top=8pt,
  bottom=8pt,
  before skip=6pt,
  after skip=6pt
]
\raggedright
\small

\textbf{Turn 1: Identify the tribe}\par\nopagebreak[4]
\textbf{Assistant Reasoning.}\par\nopagebreak[4]
\begin{lstlisting}[basicstyle=\ttfamily\small,breaklines=true,escapeinside={(*@}{@*)}]
Okay, let's break down the user's question step by step. The user is asking for the quotient of two values: the area of a reserve near a tribe with a ceremonial dance declared intangible cultural heritage, and the birth year of the scientist who discovered oxygen. ...
\end{lstlisting}

\textbf{Assistant Tool-Call.}\par\nopagebreak[4]
\begin{lstlisting}[basicstyle=\ttfamily\small,breaklines=true,escapeinside={(*@}{@*)}]
{"name": "cultural_heritage_locator", "arguments": {"heritage_type": "dance", "include_unesco": true}}
\end{lstlisting}

\textbf{Tool Response.}\par\nopagebreak[4]
\begin{lstlisting}[basicstyle=\ttfamily\small,breaklines=true,escapeinside={(*@}{@*)}]
The Maasai tribe is known for their traditional dance, which is a significant cultural heritage.
\end{lstlisting}

\turnsep

\textbf{Turn 2: Locate the reserve}\par\nopagebreak[4]
\textbf{Assistant Reasoning.}\par\nopagebreak[4]
\begin{lstlisting}[basicstyle=\ttfamily\small,breaklines=true,escapeinside={(*@}{@*)}]
Okay, so the user wants the quotient of the area of a reserve near the Maasai tribe's ceremonial dance (which is intangible cultural heritage) in hectares, divided by the birth year of the scientist who discovered oxygen. ...
\end{lstlisting}

\textbf{Assistant Tool-Call.}\par\nopagebreak[4]
\begin{lstlisting}[basicstyle=\ttfamily\small,breaklines=true,escapeinside={(*@}{@*)}]
{"name": "nearby_reserve_locator", "arguments": {"cultural_location": "Maasai tribe", "radius": 100}}
\end{lstlisting}

\textbf{Tool Response.}\par\nopagebreak[4]
\begin{lstlisting}[basicstyle=\ttfamily\small,breaklines=true,escapeinside={(*@}{@*)}]
Serengeti National Park
\end{lstlisting}

\turnsep

\textbf{Turn 3: Get reserve area}\par\nopagebreak[4]
\textbf{Assistant Reasoning.}\par\nopagebreak[4]
\begin{lstlisting}[basicstyle=\ttfamily\small,breaklines=true,escapeinside={(*@}{@*)}]
Okay, so the user wants the quotient of the area of the Serengeti National Park (near the Maasai tribe) in hectares divided by the birth year of the scientist who discovered oxygen. ...
\end{lstlisting}

\textbf{Assistant Tool-Call.}\par\nopagebreak[4]
\begin{lstlisting}[basicstyle=\ttfamily\small,breaklines=true,escapeinside={(*@}{@*)}]
{"name": "area_calculator", "arguments": {"location_name": "Serengeti National Park", "unit": "hectares"}}
\end{lstlisting}

\textbf{Tool Response.}\par\nopagebreak[4]
\begin{lstlisting}[basicstyle=\ttfamily\small,breaklines=true,escapeinside={(*@}{@*)}]
{"location_name": "Serengeti National Park", "area": 1500000, "unit": "hectares", "include_water_bodies": true, "precision_level": 3, "data_source": "satellite"}
\end{lstlisting}

\turnsep

\textbf{Turn 4: Identify oxygen discoverer}\par\nopagebreak[4]
\textbf{Assistant Reasoning.}\par\nopagebreak[4]
\begin{lstlisting}[basicstyle=\ttfamily\small,breaklines=true,escapeinside={(*@}{@*)}]
Okay, so the user wants the quotient of the area of Serengeti National Park in hectares divided by the birth year of the scientist who discovered oxygen. First, I already have the area of Serengeti National Park as 1,500,000 hectares. ...
\end{lstlisting}

\textbf{Assistant Tool-Call.}\par\nopagebreak[4]
\begin{lstlisting}[basicstyle=\ttfamily\small,breaklines=true,escapeinside={(*@}{@*)}]
{"name": "element_discovery_info", "arguments": {"element_name": "oxygen", "include_year": true}}
\end{lstlisting}

\textbf{Tool Response.}\par\nopagebreak[4]
\begin{lstlisting}[basicstyle=\ttfamily\small,breaklines=true,escapeinside={(*@}{@*)}]
{"year": 1774, "discoverer": "Joseph Priestley", "element_name": "Oxygen"}
\end{lstlisting}

\turnsep

\textbf{Turn 5: Get birth year}\par\nopagebreak[4]
\textbf{Assistant Reasoning.}\par\nopagebreak[4]
\begin{lstlisting}[basicstyle=\ttfamily\small,breaklines=true,escapeinside={(*@}{@*)}]
Okay, so the user wants the quotient of the area of Serengeti National Park in hectares divided by the birth year of the scientist who discovered oxygen. First, I already have the area of Serengeti National Park as 1,500,000 hectares. ...
\end{lstlisting}

\textbf{Assistant Tool-Call.}\par\nopagebreak[4]
\begin{lstlisting}[basicstyle=\ttfamily\small,breaklines=true,escapeinside={(*@}{@*)}]
{"name": "historical_fact_lookup", "arguments": {"person_name": "Joseph Priestley", "fact_type": "birth_year"}}
\end{lstlisting}

\textbf{Tool Response.}\par\nopagebreak[4]
\begin{lstlisting}[basicstyle=\ttfamily\small,breaklines=true,escapeinside={(*@}{@*)}]
{"result": "1733"}
\end{lstlisting}

\turnsep

\textbf{Turn 6: Final calculation}\par\nopagebreak[4]
\textbf{Assistant Reasoning.}\par\nopagebreak[4]
\begin{lstlisting}[basicstyle=\ttfamily\small,breaklines=true,escapeinside={(*@}{@*)}]
Okay, let's see. The user wants the quotient of the area of Serengeti National Park in hectares divided by the birth year of the scientist who discovered oxygen. ...
\end{lstlisting}

\textbf{Assistant Tool-Call.}\par\nopagebreak[4]
\begin{lstlisting}[basicstyle=\ttfamily\small,breaklines=true,escapeinside={(*@}{@*)}]
{"name": "advanced_calculator", "arguments": {"operation": "division", "operand1": 1500000, "operand2": 1733}}
\end{lstlisting}

\textbf{Tool Response.}\par\nopagebreak[4]
\begin{lstlisting}[basicstyle=\ttfamily\small,breaklines=true,escapeinside={(*@}{@*)}]
{"result": "865.55"}
\end{lstlisting}

\turnsep

\textbf{Turn 7: Final answer}\par\nopagebreak[4]
\textbf{Assistant Reasoning.}\par\nopagebreak[4]
\begin{lstlisting}[basicstyle=\ttfamily\small,breaklines=true,escapeinside={(*@}{@*)}]
Okay, let me go through the steps again to make sure I didn't miss anything. The user asked for the quotient of the area of Serengeti National Park in hectares divided by the birth year of the scientist who discovered oxygen. ...
\end{lstlisting}

\textbf{Assistant Final Answer.}\par\nopagebreak[4]
\begin{lstlisting}[basicstyle=\ttfamily\small,breaklines=true,escapeinside={(*@}{@*)}]
The quotient of the area of Serengeti National Park (1,500,000 hectares) divided by the birth year of Joseph Priestley (1733) is **(*@\good{865.55}@*)**.

$$
\frac{1,\!500,\!000}{1733} \approx (*@\good{865.55}@*)
$$

**Answer:** $\boxed{(*@\good{865.55}@*)}$
\end{lstlisting}

\end{tcolorbox}
\endgroup

%% file: sections/8-additional-results.tex
\section{Additional Experiments and Methodological Details}
\label{app:additional-evidence}

\subsection{Extended Training Budget}
\label{app:extended-training}

We compare SFT, MatchTIR, and \method{} on Qwen3-1.7B at 24 and 48 training updates, keeping the other settings fixed within each method. Table~\ref{tab:extended-training} reports the results for both budgets.

\begin{table}[htbp]
\centering
\small
\caption{Extended training on Qwen3-1.7B.}
\label{tab:extended-training}
\begin{tabular}{ccccc}
\toprule
Updates & Method & FTRL Solve-F1 & ToolHop AC & BFCL \\
\midrule
24 & SFT & 17.66 & 23.32 & 8.16 \\
24 & MatchTIR & 21.43 & 32.26 & 9.18 \\
24 & \method{} & 22.93 & 34.47 & 11.84 \\
\midrule
48 & SFT & 16.68 & 21.41 & 6.79\\
48 & MatchTIR & 23.42 & 38.09 & 8.38 \\
48 & \method{} & 27.17 & 40.50 & 12.62 \\
\bottomrule
\end{tabular}
\end{table}

At 48 updates, \method{} remains ahead of MatchTIR by 3.75 on FTRL Solve-F1, 2.41 on ToolHop AC, and 4.24 on BFCL. Compared with 24 updates, \method{} improves from 22.93 to 27.17 on FTRL, from 34.47 to 40.50 on ToolHop, and from 11.84 to 12.62 on BFCL. In contrast, extending SFT training reduces performance across all three benchmarks. These results show that \method{} retains and generally strengthens its advantage over the compared baselines under a larger training budget.

\subsection{Teacher Quality on a Common-Support Training Set}
\label{app:teacher-quality}

Since Qwen3-235B provides the primary source of expert traces in our main training set, we further study the effect of teacher quality in a controlled setting. We compare traces generated by Qwen3-8B and Qwen3-235B on a common-support subset. Their verified-trace acceptance rates are 55.4\% and 73.0\%, respectively. To separate trace quality from differences in accepted prompts, we form a common-support set of 1,096 training prompts for which both teachers produce a verified valid trace. All six runs in Table~\ref{tab:teacher-quality} use the same subset, a Qwen3-4B student, the same trace count, and matched training-budget and evaluation settings. Within each method, only the teacher trace source changes.

\begin{table}[htbp]
\centering
\small
\caption{Teacher-source comparison on the same training subset.}
\label{tab:teacher-quality}
\begin{tabular}{lcccc}
\toprule
& \multicolumn{2}{c}{FTRL Solve-F1} & \multicolumn{2}{c}{ToolHop AC} \\
\cmidrule(lr){2-3}\cmidrule(lr){4-5}
Method & 8B & 235B & 8B & 235B \\
\midrule
SFT & 24.66 & 26.15 & 30.46 & 31.76 \\
SFT$\rightarrow$MatchTIR & 27.72 & 28.88 & 36.21 & 37.39 \\
\method{} & \textbf{28.65} & \textbf{31.76} & \textbf{36.68} & \textbf{40.40} \\
\bottomrule
\end{tabular}
\end{table}

The stronger teacher improves \method{} by 3.11 points on FTRL and 3.72 points on ToolHop. With the 8B teacher, \method{} still has the highest scores among the three controlled methods, exceeding SFT$\rightarrow$MatchTIR by 0.93 and 0.47 points, respectively. Thus, \method{} can be trained with a much smaller teacher and still retain advantages over the matched trace-using baselines.

\subsection{Offline Data Construction and Model Training Costs}
\label{app:compute-cost}

\paragraph{Offline construction.}
Expert-trace generation, execution-based verification, and segmentation are performed before model training. Table~\ref{tab:offline-cost} reports the Qwen3-235B and Qwen3-8B construction costs. The usage costs are calculated based on OpenRouter’s pricing.

\begin{table}[htbp]
\centering
\small
\caption{Offline trace-construction API usage. Token counts are in millions; usage costs are in USD.}
\label{tab:offline-cost-235b}
\label{tab:offline-cost}
\begin{tabular}{lccc}
\toprule
Stage & Input tokens & Output tokens & Usage cost \\
\midrule
\multicolumn{4}{l}{\emph{Qwen3-235B teacher}} \\
Qwen3-235B trace generation & 62.1 & 8.97 & 44.59 \\
ChatGPT-5.4 segmentation-period calls & 10.5 & 2.84 & 55.09 \\
\textbf{235B subtotal} & \textbf{72.6} & \textbf{11.81} & \textbf{99.68} \\
\midrule
\multicolumn{4}{l}{\emph{Qwen3-8B teacher}} \\
Qwen3-8B trace generation & 44.20 & 5.72 & 7.77 \\
ChatGPT-5.4 trace segmentation & 4.84 & 1.38 & 31.60 \\
\textbf{8B subtotal} & \textbf{49.04} & \textbf{7.10} & \textbf{39.37} \\
\bottomrule
\end{tabular}
\end{table}

\paragraph{Model training.}
All measured main Qwen3-4B training jobs use four H200 GPUs. GPU-hours equal the number of allocated GPUs multiplied by the complete wall time of the successful job. For SFT$\rightarrow$MatchTIR, both stages are included. Wall times in Table~\ref{tab:student-cost} are displayed at minute resolution, whereas GPU-hours are computed from the full durations. 

\begin{table}[htbp]
\centering
\small
\caption{Measured Qwen3-4B training cost on four H200 GPUs. Offline trace construction is excluded and reported separately above.}
\label{tab:student-cost}
\begin{tabular}{lcc}
\toprule
Method & Wall time (h:min) & H200 GPU-hours \\
\midrule
MatchTIR & 14:31 & 58.11 \\
SFT$\rightarrow$MatchTIR & 01:23 + 11:46 & 52.68 \\
\method{} & 16:16 & 65.13 \\
\bottomrule
\end{tabular}
\end{table}

In these runs, \method{} uses 12.1\% more GPU-hours than MatchTIR and 23.6\% more than SFT$\rightarrow$MatchTIR. At deployment, the model receives the task prompt and its ordinary interaction history: it does not require a teacher, privileged expert traces, or additional trace-generation or segmentation calls.

\subsection{Cross-Benchmark Objective Ablation}
\label{app:cross-benchmark-objectives}

We evaluate the four Qwen3-4B objective variants from the main ablation on BFCL as well as FTRL. All four use $q=0$, disabling prompt-only anchoring, so that the comparison isolates the trace-conditioned RL surrogate and component-aware SFT. The variants are A: $\mathcal{L}^{x}_{\mathrm{RL}}$; B: $\mathcal{L}^{h}_{\mathrm{RL}}$; C: $\mathcal{L}^{x}_{\mathrm{RL}}+\beta_{\mathrm{SFT}}\mathcal{L}_{\mathrm{SFT}}$; and D: $\mathcal{L}^{h}_{\mathrm{RL}}+\beta_{\mathrm{SFT}}\mathcal{L}_{\mathrm{SFT}}$.

\begin{table}[htbp]
\centering
\small
\caption{Objective ablation on Qwen3-4B with $q=0$. BFCL Avg is the arithmetic mean of Multi-Turn, Search, and Memory accuracy.}
\label{tab:cross-benchmark-objectives}
\begin{tabular}{lrrrrrrr}
\toprule
& \multicolumn{3}{c}{FTRL} & \multicolumn{4}{c}{BFCL} \\
\cmidrule(lr){2-4}\cmidrule(lr){5-8}
Variant & Solve-R & Solve-P & Solve-F1 & Multi-Turn & Search & Memory & Avg \\
\midrule
A & 36.10 & 30.60 & 31.60 & 24.38 & \textbf{17.50} & 14.62 & 18.83 \\
B & 35.29 & 31.42 & 31.85 & 25.37 & 17.00 & 14.84 & 19.07 \\
C & 36.36 & 28.46 & 30.18 & 26.38 & 16.50 & 15.27 & 19.38 \\
D & \textbf{40.45} & \textbf{32.94} & \textbf{33.21} & \textbf{26.62} & \textbf{17.50} & \textbf{17.20} & \textbf{20.44} \\
\bottomrule
\end{tabular}
\end{table}

Variant D achieves the highest scores on all three FTRL metrics and the highest BFCL average, with the highest Multi-Turn and Memory scores and a tie with A on Search. On FTRL, adding component-aware SFT to prompt-only RL changes Solve-F1 by $-1.42$ points, whereas adding it to trace-conditioned RL improves Solve-F1 by 1.36 points. On BFCL, the corresponding gains are 0.55 and 1.37 points. These results support the empirical benefit of coupling the two losses across benchmarks.

\paragraph{Comparison under the main configuration.}
We further compare variants C and D with prompt-only anchoring
enabled ($q=0.4$).
Both use Qwen3-4B trained for 24 updates, with the same
component-aware SFT and anchoring scheme.
Under this setting, variant D corresponds to the main \method{} configuration. Table~\ref{tab:main-config-objectives} reports the results.

\begin{table}[htbp]
\centering
\small
\setlength{\tabcolsep}{5pt}
\caption{Comparison of variants C and D with prompt-only anchoring enabled ($q=0.4$).
Both variants use the same component-aware SFT.
BFCL Avg averages Multi-Turn, Search, and Memory accuracy.}
\label{tab:main-config-objectives}
\begin{tabular}{lccccccc}
\toprule
& \multicolumn{3}{c}{FTRL} & \multicolumn{4}{c}{BFCL} \\
\cmidrule(lr){2-4}\cmidrule(lr){5-8}
Variant & Solve-R & Solve-P & Solve-F1 & Multi-Turn & Search & Memory & Avg \\
\midrule
C & 36.62 & 31.16 & 31.87 & 21.50 & 10.00 & 17.85 & 16.45 \\
D (\method{}) & \textbf{42.41} & \textbf{34.69} & \textbf{36.60}
& \textbf{27.50} & \textbf{18.00} & \textbf{18.28} & \textbf{21.26} \\
\bottomrule
\end{tabular}
\end{table}

Variant D outperforms C
on all three FTRL metrics and all three BFCL subtasks,
supporting the benefit of privileged RL conditioning
under the main training configuration.

\paragraph{Controlled updates with matched and mismatched trace contexts.}
To further examine the role of task-matched expert context within joint SFT–RL training, we conduct a controlled update experiment starting from the same Qwen3-4B checkpoint obtained after 12 training iterations of \method{} with $q=0.4$. We collect a fixed batch of 256 prompts with eight prompt-only rollouts per prompt. All conditions reuse these rollouts, rewards, and advantages.

We compare three RL conditioning settings: prompt-only context (P), the corresponding expert trace (M, matched), and an expert trace from another task (X, mismatched). Mismatched traces are selected to approximately match trace length and tool sets, although tool sets are not always identical. The intervention affects only the context used for RL likelihood evaluation; the correct expert SFT targets and reward references remain unchanged. Each condition recomputes both old- and current-policy likelihoods under its own context and performs one training iteration using a freshly initialized AdamW optimizer, with the same learning rate and loss coefficients.

We evaluate both $q=0$ and $q=0.4$. For $q=0.4$, all three conditions share the same prompt-level anchoring assignment. Anchored prompts receive standard prompt-only RL without SFT; the remaining prompts use the corresponding P/M/X context with identical expert SFT inputs and supervision masks. For $q=0$, no prompts are assigned to the anchoring branch. After this single training iteration, we evaluate each updated policy on FTRL without expert traces. Table~\ref{tab:controlled-context-updates} reports the results.

\begin{table}[htbp]
\centering
\small
\caption{Controlled RL-context comparison from a shared \method{} checkpoint. All conditions use the same sampled rollouts, rewards, advantages, and correct expert SFT targets. The SFT activation rule follows the anchoring setting. Scores are percentages; bold indicates the best post-update score in each column within each $q$ setting.}
\label{tab:controlled-context-updates}
\begin{tabular}{clrrr}
\toprule
$q$ & RL Context & Solve-R & Solve-P & Solve-F1 \\
\midrule
--- & Before update & 36.41 & 30.76 & 30.89 \\
\midrule
$0$ & Prompt-only (P) & 36.98 & 29.34 & 30.52 \\
 & Mismatched trace (X) & 36.45 & 29.09 & 29.95 \\
 & Matched trace (M) & \textbf{37.21} & \textbf{31.39} & \textbf{31.22} \\
\midrule
$0.4$ & Prompt-only (P) & 36.58 & 28.29 & 29.57 \\
 & Mismatched trace (X) & 35.18 & 29.81 & 29.46 \\
 & Matched trace (M) & \textbf{38.76} & \textbf{31.17} & \textbf{31.94} \\
\bottomrule
\end{tabular}
\end{table}

Matched context achieves the highest post-update scores on all three FTRL metrics under both anchoring settings. Its Solve-F1 exceeds prompt-only context by 0.70 and 2.37 points at $q=0$ and $q=0.4$, respectively, and mismatched context by 1.27 and 2.48 points. At this checkpoint, matched context also improves all three metrics over the pre-update model. While the full-training comparisons in Tables~\ref{tab:cross-benchmark-objectives} and~\ref{tab:main-config-objectives} evaluate the overall effects of different training objectives, this experiment compares RL contexts from a shared checkpoint with rollouts, rewards, advantages, and expert SFT supervision held fixed. The results provide local evidence that task-matched expert context during RL updates can improve subsequent prompt-only performance.

\subsection{Cross-Benchmark Trace-Component Ablation}
\label{app:cross-benchmark-trace-components}

We examine the role of individual expert-trace components as privileged RL context using Qwen3-4B with prompt-only anchoring disabled ($q=0$).
Following the setup of Table~\ref{tab:hint-ablation}, we separately remove reasoning, tool-calls, or tool responses from the context used to compute the RL policy ratios.
The SFT inputs, supervision masks, and loss coefficients remain unchanged across variants, so the comparison concerns the role of each component as RL context rather than as a direct SFT target.
Table~\ref{tab:cross-benchmark-trace-components} presents the detailed results.

\begin{table}[htbp]
\centering
\small
\setlength{\tabcolsep}{4pt}
\caption{Privileged trace-component ablation using Qwen3-4B on FTRL and BFCL ($q=0$). RE, TC, and TR denote reasoning, tool-calls, and tool responses. \xmark{} marks the removed component; SFT remains unchanged. BFCL Avg is the arithmetic mean of Multi-Turn, Search, and Memory accuracy.}
\label{tab:cross-benchmark-trace-components}
\begin{tabular}{ccc|ccc|cccc}
\toprule
\multicolumn{3}{c|}{} & \multicolumn{3}{c|}{FTRL} & \multicolumn{4}{c}{BFCL} \\
\cmidrule(lr){4-6}\cmidrule(lr){7-10}
RE & TC & TR & Solve-R & Solve-P & Solve-F1 & Multi-Turn & Search & Memory & Avg \\
\midrule
\xmark & & & 35.31 & 27.46 & 26.12 & 26.12 & 14.00 & 13.55 & 17.89 \\
 & \xmark & & 39.19 & 28.52 & 31.11 & 24.62 & 14.00 & 10.32 & 16.31 \\
 & & \xmark & 31.32 & 25.27 & 25.94 & 23.75 & 10.50 & 10.54 & 14.93 \\
\midrule
\multicolumn{3}{c|}{All} & \textbf{40.45} & \textbf{32.94} & \textbf{33.21} & \textbf{26.62} & \textbf{17.50} & \textbf{17.20} & \textbf{20.44} \\
\bottomrule
\end{tabular}
\end{table}

On FTRL, removing reasoning or tool responses lowers Solve-F1 from 33.21 to 26.12 and 25.94, respectively, while removing tool-calls lowers it to 31.11.
The larger drops from removing reasoning and tool responses highlight the value of expert reasoning and environmental feedback as context for policy optimization.
The drop from removing tool-calls shows that expert action information remains useful when tool-call SFT supervision is retained.
Using the complete trace achieves the highest scores on all three FTRL metrics among the compared variants.

\subsection{More Choices of SFT Schedule}
\label{app:more_sft_schedule}

We compare SFT schedules with Qwen3-4B on FTRL, keeping trace-conditioned RL fixed and prompt-only anchoring disabled ($q=0$). Table~\ref{tab:sft_schedule} compares schedules within each component, while Table~\ref{tab:more_sft_schedule} includes single-component supervision and joint token-count annealing.

\begin{table}[htbp]
\centering
\small
\caption{Additional comparison of SFT schedule designs using Qwen3-4B on FTRL. ``Seg A'' denotes annealing over semantic reasoning segments, ``Scale A'' denotes annealing the tool-call loss scale, and ``Full Token A'' denotes token-count annealing over concatenated reasoning and tool-call tokens.}
\begin{tabular}{cc|ccc}
\toprule
\textbf{Reasoning} & \textbf{Tool-Call} & \textbf{Solve-R} & \textbf{Solve-P} & \textbf{Solve-F1} \\
\midrule
Seg A & None & 35.11 & 32.67 & 31.22 \\
None & Scale A & 35.01 & 25.51 & 27.69 \\
\multicolumn{2}{c|}{Full Token A} & 38.76 & 29.65 & 31.91 \\
\midrule
\textbf{Seg A} & \textbf{Scale A} & \textbf{40.45} & \textbf{32.94} & \textbf{33.21} \\
\bottomrule
\end{tabular}
\label{tab:more_sft_schedule}
\end{table}

For reasoning, \textit{Fix} supervises the $L_4$ prefix retained by preprocessing in each assistant turn throughout training, with a loss scale of 1.0. For tool-calls, \textit{Fix} supervises complete spans with $\lambda_a=1.0$. \textit{Seg A} and \textit{Scale A} follow the semantic-prefix and tool-call weight schedules in Section~\ref{sec:method-component-sft}. For the two \textit{Token A} variants in Table~\ref{tab:sft_schedule}, a target component of $N$ tokens within an assistant turn retains direct supervision on its first
\begin{align}
N_{\mathrm{sup}}(s)=\left\lfloor N\left(1-\frac{s-1}{24}\right)\right\rfloor,
\qquad s=1,\ldots,24, 
\end{align}
tokens. This mask applies to reasoning text in \textit{Token A / Scale A}, and to the concatenated tool-call text within each turn in \textit{Seg A / Token A}; the latter assigns selected tool-call tokens a fixed weight of 1.0. Unselected tokens remain in the input context. The retained fraction decreases from 1 to $1/24$ before rounding. \textit{None} omits the corresponding SFT component. The outer SFT coefficient remains $\beta_{\mathrm{SFT}}=0.1$.

The full component-aware schedule achieves higher Solve-F1 than reasoning-only or tool-call-only supervision (33.21 versus 31.22 and 27.69). \textit{Full Token A}, which anneals supervision over concatenated reasoning and tool-call tokens, improves Solve-F1 over reasoning-only supervision to 31.91, while reducing Solve-P from 32.67 to 29.65. The component-aware schedule achieves the highest scores on all three metrics in Table~\ref{tab:more_sft_schedule}, supporting separate schedules for reasoning and tool-calls.

\section{Limitations}
\label{app:limitations}

This work has several limitations. First, PACT relies on expert tool-use traces during training. These traces are used only as privileged optimization signals and are not needed at inference time, but their construction, verification, and segmentation introduce additional offline data-preparation costs. Second, our current task focuses on agent--tool interactions rather than ongoing dialogue with users. Extending PACT to settings where agents seek clarification and adapt their tool use to user feedback is a promising direction for future work. Finally, PACT uses predefined schedules for reasoning-prefix supervision and tool-call loss weighting, together with a fixed prompt-only anchoring probability. Adapting these choices to training dynamics remains an avenue for future work.